\documentclass[sigconf]{acmart}

\usepackage{color}
\usepackage{colortbl}
\usepackage[flushleft]{threeparttable}
\usepackage{amsmath,amsfonts,amssymb}
\usepackage{tikz}
\usepackage{xspace}
\usepackage{bbding} 
\usepackage{romannum}
\usepackage{enumitem}

\AtBeginDocument{%
  \providecommand\BibTeX{{%
    \normalfont B\kern-0.5em{\scshape i\kern-0.25em b}\kern-0.8em\TeX}}}

\copyrightyear{2020} 
\acmYear{2020} 
\setcopyright{acmlicensed}\acmConference[KDD '20]{Proceedings of the 26th ACM SIGKDD Conference on Knowledge Discovery and Data Mining}{August 23--27, 2020}{Virtual Event, CA, USA}
\acmBooktitle{Proceedings of the 26th ACM SIGKDD Conference on Knowledge Discovery and Data Mining (KDD '20), August 23--27, 2020, Virtual Event, CA, USA}
\acmPrice{15.00}
\acmDOI{10.1145/3394486.3403077}
\acmISBN{978-1-4503-7998-4/20/08}

\begin{document}
\fancyhead{}
\title{Laplacian Change Point Detection for Dynamic Graphs}
\newcommand{\method}{LAD\xspace}


\author{Shenyang Huang}
\email{shenyang.huang@mail.mcgill.ca}
\affiliation{%
  \institution{Mila, McGill University}
}

\author{Yasmeen Hitti}
\email{yasmeen.hitti@mail.mcgill.ca}
\affiliation{%
  \institution{Mila, McGill University}
}

\author{Guillaume Rabusseau}
\email{guillaume.rabusseau@umontreal.ca}
\affiliation{%
  \institution{Mila \& DIRO, Universit\'e de Montr\'eal - CIFAR AI chair}
}

\author{Reihaneh Rabbany}
\email{rrabba@cs.mcgill.ca}
\affiliation{%
  \institution{Mila, McGill University - CIFAR AI chair}
}



\renewcommand{\shortauthors}{Huang,S.; Hitti,Y.; Rabusseau,G.; Rabbany,R.}

\begin{abstract}
    Dynamic and temporal graphs are rich data structures that are used to model complex relationships between entities over time. In particular, anomaly detection in temporal graphs is crucial for many real world applications such as intrusion identification in network systems, detection of ecosystem disturbances and detection of epidemic outbreaks. In this paper, we focus on change point detection in dynamic graphs and address two main challenges associated with this problem: \Romannum{1}) how to compare graph snapshots across time, \Romannum{2}) how to capture temporal dependencies. To solve the above challenges, we propose Laplacian Anomaly Detection~(LAD) which uses the spectrum of the Laplacian matrix of the graph structure at each snapshot to obtain low dimensional embeddings. LAD explicitly models short term and long term dependencies by applying two sliding windows. In synthetic experiments, LAD outperforms the state-of-the-art method. We also evaluate our method on three real dynamic networks: UCI message network, US senate co-sponsorship network and Canadian bill voting network. In all three datasets, we demonstrate that our method can more effectively identify anomalous time points according to significant real world events.
\end{abstract}

\maketitle

\section{Introduction and Motivation} 
\label{Sec:intro}

Real world problems in various domains~(e.g. political science, biology, chemistry and sociology) can be modeled as evolving networks that capture temporal relations between nodes. With the increasing availability of dynamic network data in areas such as social media, public health and transportation, providing sophisticated methods that can identify anomalies over time is an important research direction. The goal of anomaly detection in dynamic graphs is to identify different types of time-varying anomalies that significantly deviate from the "normal" or "expected" behavior of the underlying graph distribution. We provide some motivating examples below.

In a transportation network, traffic volumes can be represented as edge weights in isolated network snapshots that are taken at various times of the day. When a traffic accident occurs, a drastic change in that section of the network will likely follow~(e.g. decreasing amount of traffic). On the other hand, cyberbullying, terrorist attack planning and fraud information dissemination~\cite{yu2016survey} can all be seen as cases of anomalies in a temporal social graph. Identifying these anomalies accurately and rapidly can result in positive and significant real world impact. Lastly, complex clinical information can also be represented as a dynamic graph. Patients, symptoms and treatments can be represented as vertices in a heterogenous network. Detecting anomalies in such network can discover critical scenarios where errors or abnormal patient conditions arise~\cite{chen2012community}.

\begin{figure}[t]
    \begin{center}
        \centerline{\includegraphics[width=\columnwidth]{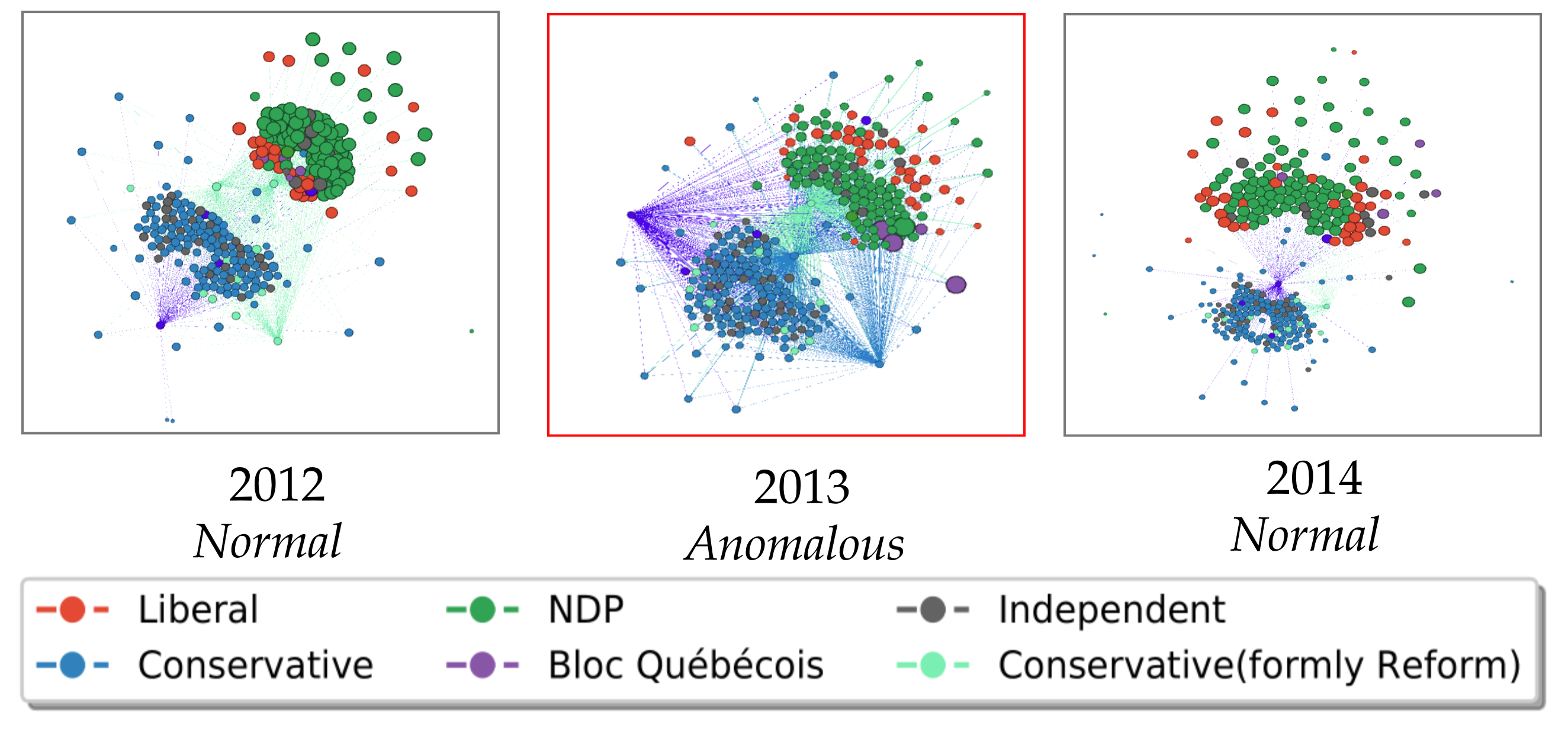}}
        \caption{\method detects changes in Canadian Member of Parliament voting patterns. \method identified 2013 as anomalous due to abnormal amount of edges between political parties.}
        \label{Fig:compare}
    \end{center}
    \vskip -0.3in
\end{figure}

\begin{table*}[t]
\centering
\begin{tabular}{c|c c c c c c}
\hline
\hline
Method & event & change point & scalable & evolving \# nodes & node permutation invariant\\
\hline
\rowcolor{gray!20}\method~(ours) & \CheckmarkBold  & \CheckmarkBold  & \CheckmarkBold  & \CheckmarkBold  & \CheckmarkBold  \\
Activity vector~\cite{ide2004eigenspace} & \CheckmarkBold  &  & &\CheckmarkBold &\CheckmarkBold \\
TENSORSPLAT~\cite{koutra2012tensorsplat} & \CheckmarkBold  & \CheckmarkBold  & & &\CheckmarkBold  \\
EdgeMonitoring~\cite{wang2017fast} & & \CheckmarkBold  & \CheckmarkBold & & \\
\hline\hline
\end{tabular}
\caption{\method satisfies all the desired properties while alternative methods lack one more.}
\label{tab:features}
\vskip -0.3in
\end{table*}

In this work, we focus on change point detection which identifies time steps where the graph structure or components deviate significantly from the normal behavior. As change point detection and event detection are closely related, we first explain the distinctions between them. Following~\cite{wang2017fast}'s definition, a \emph{change point} is a time point where there is a sudden change in the underlying network generative process and this new process continues beyond the current point. In contrast, an \emph{event} is defined as a time point where the network deviates significantly from the expected behavior and falls back to normal after this point. Once the particular anomalous graph instance is found, potential causes can be then identified through various static graph analysis techniques. Figure~\ref{Fig:compare} shows the anomalous snapshot~(inside the red box) detected by our method in the Canadian parliament voting network. Nodes are colored by political parties and node sizes are weighted using PageRank~\cite{page1999pagerank}. Our method detected anomalous cross party interaction in 2013.

There are two major challenges for change point detection:\\
\emph{1). How to compare graph snapshots across time.} To compare graph level changes over time, a summary of the network in the form of a low dimensional representation is often used~\cite{yu2016survey}. We propose to use the singular values of the Laplacian matrix~(Laplacian spectrum) because it represents a global view of the graph snapshot and of its connection to graph connectivity and low rank approximation. Due to these connections, we show that \method can detect a wide range of graph changes~(i.e. community structure and average edge weight) in dynamic graphs in Section~\ref{tab:results}.\\
\emph{2). How to capture temporal dependencies.} In practice, graphs can undergo abrupt changes at any given time step~\cite{gahrooei2018change}. These sudden changes can be identified through a short term sliding window~\cite{akoglu2010event, ide2004eigenspace}. However, points in time that signal a change in graph pattern for a long duration of time can also have high significance~\cite{ranshous2015anomaly}. In particular, it is possible for the underlying graph generation model to evolve~\cite{peel2015detecting,wang2017fast}. Effective detection of these points of change would require the model to reason with the graph behaviors beyond the most recent ones. We propose to use two context windows which explicitly compare the current graph structure with the typical behaviors from both short term and long term perspectives.\\
\textbf{Summary of contributions}: 
\begin{itemize}[topsep=0pt]
    \item We introduce a novel change point detection method : Laplacian Anomaly Detection~(LAD). LAD computes the Singular Value Decomposition~(SVD) of the graph Laplacian to obtain a low dimension graph representation. To the best of our knowledge, this is the first time that Laplacian spectrum has been used for change point detection.
    \item \method explicitly captures both the short term and the long term temporal relations to model the abrupt and gradual changes in dynamic networks. 
    \item We extensively evaluate \method on two synthetic experiments and three real world datasets. We show that \method is more effective at identifying significant events than state-of-the-art methods. We also interpret the predicted anomaly scores by looking at its correlation with different graph properties. 
\end{itemize}
\textbf{Reproducibility}: code and data is publicly available~\footnote{\url{https://github.com/shenyangHuang/LAD}}.

\section{Related Work}
\label{Sec:related}
In 2015, Ranshous et al.~\cite{ranshous2015anomaly} classified methods for anomaly detection in dynamic graphs into five categories: community based, compression based, decomposition based, distance based and probabilistic model based methods. The common strategy across all these methods are to extract a low dimensional representation from graph snapshots and then apply an anomaly scoring function to compare these representations. Most methods model temporal patterns by either using a decay function to put more emphasis on recent graphs or a manually defined sliding window~(see e.g., ~\cite{li2009temporal,miller2012scalable,thompson2009dapa}). In this section, we discuss related literatures from event detection and change point detection. In addition, we summarize features of \method and other alternative methods in Table~\ref{tab:features}. Note that \method is the only approach that satisfies all the desired properties. 

\subsection{Event Detection}
An early work by Idé and Kashima~\cite{ide2004eigenspace} aimed to find time points where the majority of the edge attributes in the network show significant deviation from the recent ones. The principal eigenvector corresponding to the maximum eigenvalue of the positive weighted adjacency matrix $W$ was used as a low dimensional representation of the graph~(called \emph{activity vector}). The typical graph behavior within a short term context window is summarized as the principle left singular vector (of the matrix formed by activity vectors in this window). The deviation of the current activity vector from the typical behavior was used as the anomaly score. Different from Idé and Kashima, we use the Laplacian spectrum to summarize graph structures and explicitly constructs two sliding windows for long term and short term context. 

Despite having a similar workflow to~\cite{ide2004eigenspace}, Akoglu and Faloutsos~\cite{akoglu2010event} focused on node embedding and change points where many nodes deviate from their normal 'behavior'. This is achieved by extracting the time sequence of selected network features for all nodes. Then, a correlation matrix representing pairwise node interactions are used the graph summary instead. Taking inspiration from Akoglu and Faloutsos, we instead suggest to incorporate outliers from the univariate temporal sequences of different network properties to understand the correlation between such properties and explaining the anomalies detected by \method in a more interpretable way. 

Koutra et al.~\cite{koutra2012tensorsplat} first formulated dynamic graphs as high order tensors and proposed to use the PARAFAC decomposition~\cite{bro1997parafac,harshman1970foundations} to obtain vector representations for anomaly scoring. The proposed TENSORSPLAT method is included as one of the alternative methods in Section~\ref{Sec:dataset}.  For more information on tensor decompositions, we refer the readers to~\cite{kolda2009tensor}. 
Shah et al. ~\cite{shah2015timecrunch} proposed an effective algorithm TimeCrunch for dynamic graph summarization using Minimum Description Length (MDL), where temporal structures are transmitted through adjacency tensors to the model. The idea behind this algorithm is to find the smallest model that will cut down the encoding length for one-shot, ranged, periodic and flickering structures in the dynamic graph. 


\subsection{Change Point Detection}
\label{Sec:def}
Koutra et al.~\cite{koutra2016deltacon} formally stated the axioms and desired properties of functions that measure the connectivity difference between two graphs~(graph similarity functions). DeltaCon computes pairwise node affinities in the first graph and then measures the difference in node affinity score of the two graphs. However, DeltaCon is only well-defined for pairs of graphs thus lacking the ability to reason with a sequence of graph snapshots. This inherently limits its ability to detect gradual changes which involve multiple snapshots. 

Peel and Clauset~\cite{peel2015detecting} first formalized the change point detection problem as identifying the times at which the large-scale patterns of interaction change fundamentally. Their proposed LetoChange method relies on an appropriate choice of a parametric family of probability distribution which describes the data. Then, a Bayesian hypothesis test is used to accept or reject if a parameter change has occurred in the model. 

Recently, Wang et al.~\cite{wang2017fast} model network evolution as a first order Markov process thus deriving their EdgeMonitoring method based on MCMC sampling theory. Their assumption is that there is some unknown underlying model that governs the generative process. Moreover, each graph snapshot is dependent on the current generative model as well as the previously observed snapshot. This method is often regarded as the current state-of-the-art for change point detection. However, EdgeMonitoring relies on consistent node orderings across all time steps. In addition, EdgeMonitoring assumes constant number of nodes for each snapshot. This assumption is easily violated in large social networks where users accounts are added frequently. In contrast, \method can manage varying number of nodes across time. 

Eswaran et al.~\cite{eswaran2018spotlight} proposed SPOTLIGHT to detect anomalous graph snapshots involving the sudden appearance or disappearance of large dense subgraphs. The core idea is to compose a $K$-dimensional sketch containing $K$ subgraphs to detect changes in the dynamic graph. While SPOTLIGHT focuses on dense subgraphs, LAD detect more variety of change points such as changes in the community structure. 

\section{Problem definition}

\subsection{Dynamic Graph} Let the interval of interest be from timestamp $1$ to $T$. A corresponding set of graph snapshots $\mathbf{G}$ is written as $\{ \mathcal{G}_t \}_{t=1}^{T}$, where each $\mathcal{G}_t = (\mathcal{V}_t, \mathcal{E}_t)$ represents the static graph at timestamp t. $\mathcal{V}_t$ and $\mathcal{E}_t$ are the set of nodes and edges respectively. Define an edge $e=(i,j,w) \in \mathcal{E}_t$ as the connection between node $i$ and node $j$ at timestamp $t$ in the dynamic graph with weight $w$. Note that as each edge $e$ is defined for a particular timestamp $t$, edges can disappear or reappear in the dynamic graph across different timestamps. By convention, $w=1$ for all edges in unweighted graphs and $w \in \mathbb{R}^{+}$ for weighted graphs. We use an adjacency matrix $\mathcal{A}_t \in \mathbb{R}^{n \times n}$ to represent edges in $\mathcal{E}_t$ where $n = |\mathcal{V}_t|$. Similar to~\cite{zheng2019addgraph, yu2018netwalk, ide2004eigenspace, koutra2012tensorsplat}, the number of nodes in the graph is assumed to be constant across all timestamps (thus maintaining the shape of the adjacency matrix $\mathcal{A}_t$). However, our method is also applicable to real world networks where the number of active nodes fluctuates from one snapshot to another. 
\begin{figure}[t]
    \begin{center}
        \centerline{\includegraphics[width=\columnwidth]{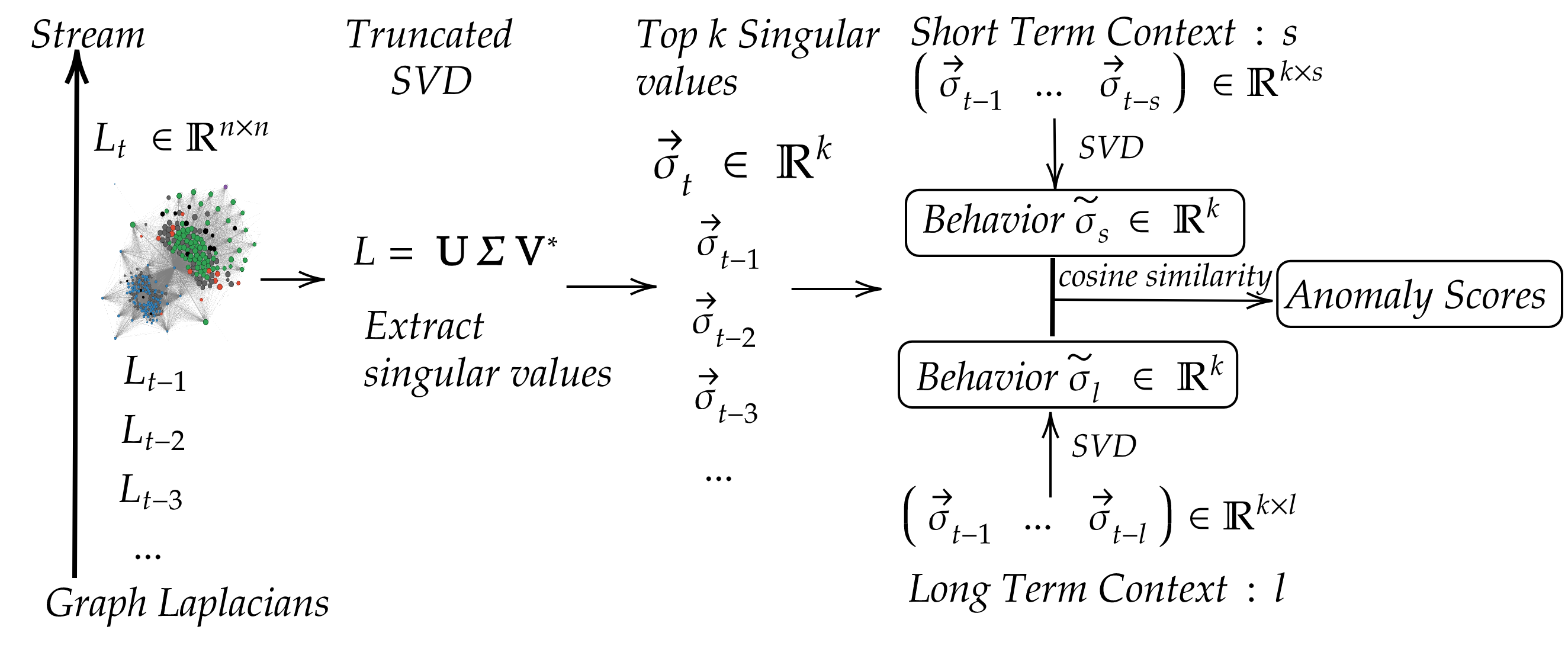}} \vskip -0.1in
        \caption{\method considers short and long term temporal relations encoded in the graph Laplacian spectrum. }        
        \label{Fig:framework}
    \end{center}
    \vskip -0.3in
\end{figure}
\subsection{Change Point Detection} Based on the above formulation, the goal is to find anomalous graphs $\mathcal{G}_t$ in $\mathbf{G}$. Given an anomaly scoring function $f: \mathcal{G}_t \rightarrow \mathbb{R}$, find time steps t such that $|f(\mathcal{G}_t) -  f(\mathcal{G}_N)| > \delta$ or $|f(\mathcal{G}_t) -  f(\mathcal{G}_W)| > \epsilon$ where $\mathcal{G}_N$ is the normal behavior of the graph in the global context, $\mathcal{G}_W$ is the short term behavior of the graph in recent context window $W$ and $\delta, \epsilon$ are thresholds. In general, the anomaly scoring function should clearly differentiate anomalous points from normal ones and assign higher anomaly scores to more anomalous points. 

\section{Laplacian Anomaly Detection}
\label{Sec:method}
We propose a new spectral anomaly detection method for dynamic graphs:  Laplacian Anomaly Detection~(LAD).
The core idea of \method is to detect high level graph changes from low dimensional embeddings~(called signature vectors). The "typical" or "normal" behavior of the graph can be extracted from a stream of signature vectors based on both short term and long term dependencies. In this way, we can compare the deviation of current signature vector from the normal behavior. Figure~\ref{Fig:framework} shows the flowchart of our method.

\subsection{Laplacian Spectrum}
\begin{figure}[t]
    \begin{center}
        \centerline{\includegraphics[width=\columnwidth]{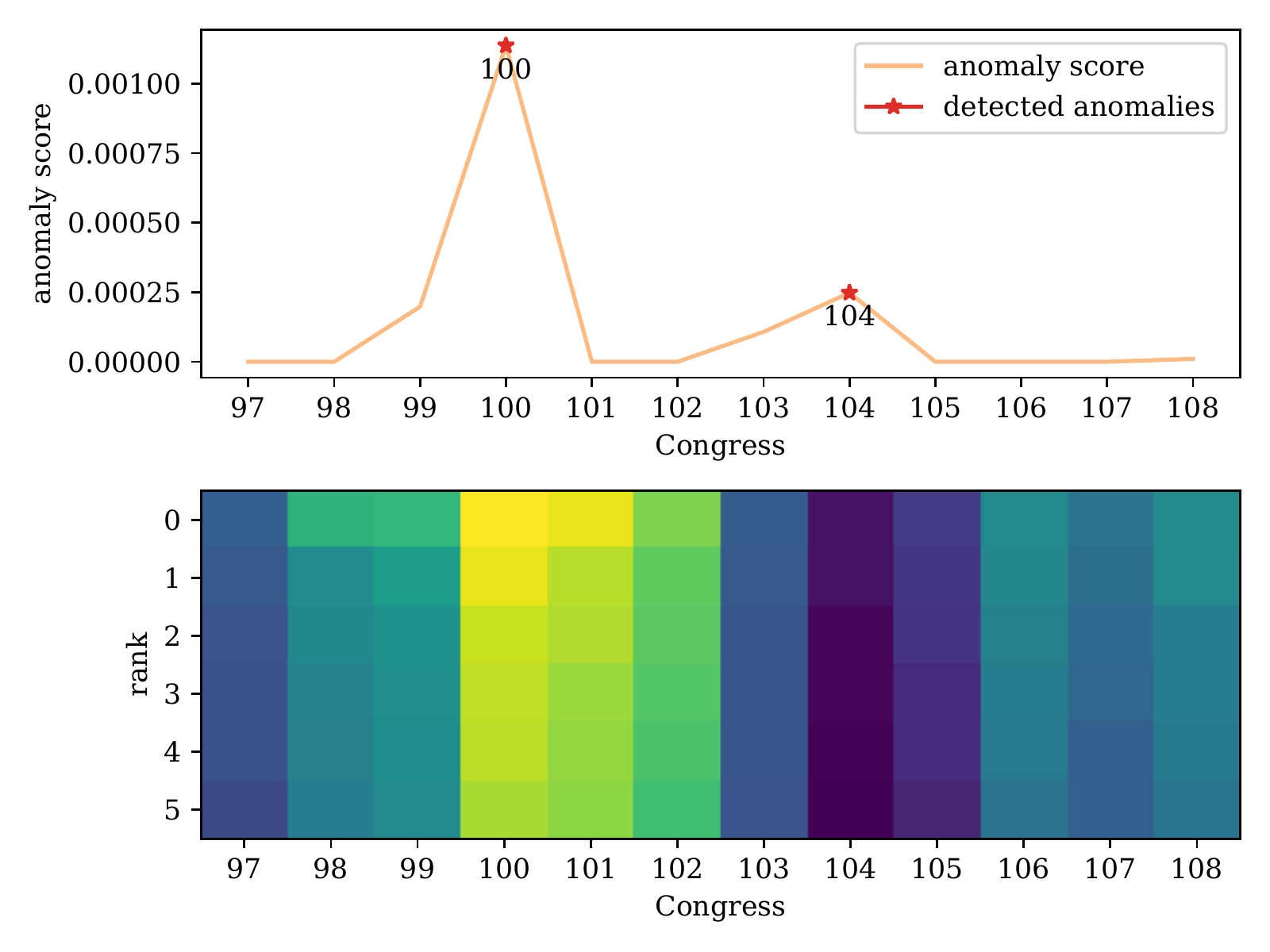}}
 \vskip -0.2in        \caption{\method captures changes in the graph spectrum. \method scores (top) and the top 6 singular values (bottom) are aligned and show correspondence at each time step. This illustration is from the Senate co-sponsorship network. The warmer the color, the higher the intensity.}        
        \label{Fig:spectro}
    \end{center}
    \vskip -0.3in
\end{figure}

We define the~(unnormalized) Laplacian matrix $\mathbf{L}_t$ as $\mathbf{L}_t = \mathbf{D}_t - \mathbf{A}_t$ where $\mathbf{D}_t$ is the diagonal degree matrix and $\mathbf{A}_t$ is the adjacency matrix of $\mathcal{G}_t$. In this work, we choose the singular values obtained through Singular Value Decomposition~(SVD)~\cite{golub1971singular} of the Laplacian matrix as graph embeddings for each snapshot. Figure~\ref{Fig:spectro} shows the visualization of the Laplacian spectrum and the corresponding anomaly scores detected by \method for the Senate co-sponsorship network. This choice is motivated by the following. \\
\textbf{Singular values:} 
\begin{itemize}
    \item are related to the Laplacian spectrum,
    \item encodes the compression loss of low rank approximations of the Laplacian matrix,
    \item are node permutation invariant,
    \item can be efficiently computed in real world sparse matrices
\end{itemize}

First, it is known that the singular values of a positive symmetric matrix coincides with its eigenvalues. The Laplacian matrix is symmetric and positive semi-definite for an undirected weighted graph~\cite{von2007tutorial}. The eigenvalues of the Laplacian matrix capture fundamental structural properties of the corresponding graph. This property has been extensively leveraged in many fields such as randomized algorithms, combinatorial optimization problems and machine learning~\cite{zhang2011laplacian}, and the field of spectral graph theory~\cite{chung1997spectral} is dedicated to the study of graph Laplacian matrices. As an illustration, the multiplicity $k$ of the eigenvalue $0$ of $\mathbf{L}$ equals the number of connected components in the graph~\cite{von2007tutorial}. In addition, the Laplacian spectrum is related to many other structural properties of the graph such as the degree sequence, number of connected components, diameter, vertex connectivity and more~\cite{zhang2011laplacian}. 

Second, it is well known that the truncated SVD gives the best low rank approximation of a matrix with respect to both the Frobenius norm and the 2-norm (see Eckart-Young theorem). More precisely, the $(k+1)$th singular value $\sigma_{k+1}$ of a matrix corresponds to the reconstruction error of the best rank $k$ approximation measured in the 2-norm. From this insight, we know that the~(ordered) singular spectrum $\sigma_1, \sigma_2, ..., \sigma_r$ encodes rich information regarding the reconstruction loss that would occur for different levels of low rank approximations. Truncated SVD is often used as a powerful compression tool for images~\cite{rufai2014lossy}, videos~\cite{benjamin2017compressed} and audios~\cite{zamani2017frequency}. Thus, capturing the singular spectrum can be seen an alternative compression based anomaly detection technique as categorized by Ranshous et al.~\cite{ranshous2015anomaly}. Intuitively, huge fluctuations in the singular values of the Laplacian matrix reflect drastic changes to the global graph structure. In an undirected graph where the Laplacian singular values coincides with the eigenvalues, the decrease in the number of zero singular values would reflect the decrease in the number of connected components. 

Third, change point detection can be viewed as a binary graph classification problem. Being node permutation invariant is one of the most desirable properties for a graph learning method~\cite{xu2018powerful}. However, preserving the same node ordering for each graph snapshot might not be feasible in many practical settings. Since applying row or column permutations on the Laplacian matrix has no effect on the singular values,  \method is node permutation invariant. In this way, \method can be applied to a broader range of scenarios than methods relying on a consistent node ordering such as EdgeMonitoring~\cite{wang2017fast}.

Lastly, singular values can be efficiently obtained through sparse computations~\cite{aharon2006k}. As many real world graphs are sparse, the proposed graph embedding can be scaled to large datasets. In addition, depending on computational budget, one can compute the top $k$ singular values through truncated SVD with much less computational cost when compared to the full SVD. In Section~\ref{Sec:dataset}, we show that for some real world datasets, it is often enough to compute the top $k$ singular values. We also explain the computational complexity of \method  in Section~\ref{Sub:CC}. 

\subsection{Characterizing Normal Behavior}
\label{Sub:normal}
Identifying a normal or typical behavior from a temporal window is often an integral part of change detection. Similar to~\cite{akoglu2010event, ide2004eigenspace}, we compute a "typical" or "normal" behavior vector from the previous $l$ singular spectrums where $l$ is the sliding window size. First, we perform \emph{L2} normalization on the Laplacian spectrums seen so far $\vec{\sigma}_0, \hdots, \vec{\sigma}_t$ to obtain unit vectors. Next, a context matrix $\mathbf{C}$ is constructed:
\begin{equation}
    \mathbf{C} = \left( \begin{array}{cccc}
                 |& | & &|  \\
                 \vec{\sigma}_{t-l-1} & \vec{\sigma}_{t-l-2} & \hdots & \vec{\sigma}_{t-1} \\ 
                 |& | & & |
            \end{array} \right) \in \mathbb{R}^{n \times l}
\end{equation}
where $n$ is length of the signature vector. We compute the left singular vector of $\mathbf{C}$ with SVD to obtain the normal behavior vector $\tilde{\sigma}_t$. In literature, this is often considered as a weighted average vector from the sliding window~\cite{akoglu2010event}. 

\subsection{Two Perspectives}
Different from~\cite{akoglu2010event, ide2004eigenspace}, we propose to compare the current signature vector with the typical behavior from two independent sliding windows: a short term window and a long term window. The short term window encodes information from the most recent trend and captures abrupt changes in the overall graph structure. Depending on the application, the length of the short term window can be adjusted to best reflect an appropriate time scale. On the other hand, a long term window is designed to capture larger scale and more gradual trends in the dynamic graph. For example, for the UCI Message dataset, the short term context can be monitoring weekly change and the long term context can be monitoring a biweekly change. 

\subsection{Anomalous Score Computation}
After capturing the normal behavior, one can then define a scoring function to measure the difference between the current signature vector and the expected or normal one. In this work, we use the same anomaly score as introduced in~\cite{ide2004eigenspace, akoglu2010event}, namely the $Z$ score. Let $\tilde{\sigma}_t$ be the normal behavior vector and $\vec{\sigma}_t$ be the Laplacian spectrum at current step. As mentioned in Section~\ref{Sub:normal}, both $\tilde{\sigma}_t$ and $\vec{\sigma}_t$ are normalized to unit vectors, then the $Z$ score is computed as:
\begin{equation}
\label{Eq:Zscore}
    Z = 1 - \frac{\vec{\sigma}_t^\top \tilde{\sigma}_t }{ \|\vec{\sigma}_t\|_2 \|\tilde{\sigma}_t\|_2} = 1 - \vec{\sigma}_t^\top \tilde{\sigma}_t = 1 - \cos{\theta},
\end{equation}
where $\cos{\theta}$ is the cosine similarity between $\vec{u}$ and $\vec{v}$. Essentially the $Z$ scores becomes closer to 1 when the current spectrum is very dissimilar to the norm thus signaling an anomalous point. 

Let $s$ and $l$ be the sizes of the short term and long term sliding windows. Analogous to Section~\ref{Sub:normal}, one can obtain two different normal behavior vectors $\tilde{\sigma}_s$ and $\tilde{\sigma}_l$. $\tilde{\sigma}_s$ encodes the expected norm within a few time steps while $\tilde{\sigma}_l$ captures the normal network activity within a larger time span. One can then compute the short term and long term anomaly scores $Z_{s}$ and $Z_l$ based on cosine similarity. To best aggregate these two perspectives, we take  $Z_t = max(Z_s, Z_l)$ to decide if the current graph is more anomalous in abrupt or gradual changes. 

Now having a sequence of anomaly scores $Z_1, \hdots, Z_t$, how to best select the change points based on these scores? Different than~\cite{ide2004eigenspace, akoglu2010event}, we choose the points that have the largest increase in anomaly score when compared to the previous time step. Therefore, we have the final anomaly score $Z_t^* = min(Z_t - Z_{t-1},0)$. The points with the largest $Z^*$ are then selected as anomalies.

\subsection{Computational Complexity}
\label{Sub:CC}
Given a matrix $\mathbf{A} \in \mathbb{R}^{m \times n}$, the computational complexity for full SVD is $\mathcal{O}(m^2n)$ when $m\leq n$. For a truncated rank $k$ SVD, the complexity is $\mathcal{O}(mnk)$. Halko et al.~\cite{halko2011finding} showed that randomization offers a powerful tool for performing low-rank matrix approximation. Randomized SVD more efficiently utilizes computational architectures thus achieving $\mathcal{O}(mn\log(k))$ cost. At the same time, Berry~\cite{berry1992large} presented strong numerical methods for computing SVD in large sparse matrices on a multiprocessor architecture. As many real world networks are sparse, sparse SVD can achieve significant computational savings. In this work, we use the Scipy~\cite{2020SciPy-NMeth} sparse SVD implementation in python.

\section{Experiments}
\label{Sec:dataset}
In this section, we perform extensive experiments on different types of synthetic and real world datasets to validate the effectiveness of the \method framework.

\subsection{Measuring Performance}
For quantitative evaluation of a change point detection method, we use Hits at $n$~($H@n$) metric which reports the number of identified significant anomalies out of the top $n$ most anomalous points.
In synthetic experiments, we use the ground truth labels in the generation process for evaluation. 
In real world datasets, we treat the well-known anomalous times steps as ground truth anomalies which should be detected by a given method.

\subsection{Contenders and Baselines}
We compare \method with the following alternative methods. The same short term and long term window sizes~($s$ and $l$) are used if applicable. The short term and long term anomaly scores are both shown in Figure~\ref{Fig:SBMscore},\ref{Fig:eventChange},\ref{Fig:UCIscore},\ref{Fig:senateScore},\ref{Fig:canVotescore}.
Note that for all experiments, the startup period~($0,\hdots,l$) is set to have anomaly score of 0 because we assume these points are not change points.  
\begin{itemize}[leftmargin=12pt]
    \item \textbf{Activity vector.} We refer to the method proposed by Idé et al.~\cite{ide2004eigenspace} as "activity vector" based methods. Idé et al. used the principal eigenvector of the adjacency matrix~(namely the activity vector) instead of our proposed Laplacian spectrum. According to the original work, only a short term context window is considered. 
    
    \item \textbf{TENSORSPLAT.} Koutra et al.~\cite{koutra2012tensorsplat} proposed to view the temporal graph as a tensor and then perform PARAFAC decomposition to obtain low dimensional factors that groups similar entities or timestamps together~(we use CP rank of 30 for all experiments). The original paper proposed to use clustering on the factors for change detection. However, the clustering algorithm is not specified. We use the well-known Local Outlier Factor~(LOF)~\cite{breunig2000lof} approach along with the TENSORSPLAT framework.
    
    \item \textbf{EdgeMonitoring.} Wang et al.~\cite{wang2017fast} proposed the EdgeMonitoring approach and used joint edge probabilities as the "feature vector" while modelling network evolution as a first order Markov process. 
\end{itemize}

\subsection{Interpreting Results}
It is often difficult to interpret the anomaly score of a given change point detection method as the task inherently demands direct comparison between global graph structures over time. As network characteristics vary drastically across domains~\cite{broido2019scale}, it is important to design metrics that help us understand the correlation between  anomaly scores and well-known graph properties. In this work,we identify temporal outliers in specific graph properties and compare them to the ones predicted by \method . We choose the outlier score $y$ as follows:
\begin{equation}
    y = \frac{|\alpha_t - \alpha_{avg} |}{\alpha_{std}},
\end{equation}
where $\alpha_{avg}$, $\alpha_{std}$ are the average and standard deviation of $\alpha$ computed from a moving window. We select the moving window size to correspond with the short term window size $s$. Then we compute the Spearman rank correlation~\cite{ziegel2001standard} to understand the statistical dependence between two ranked variables. We observe that \method is not relying on one particular graph statistics but rather on the most important aspects of the dynamic graph of interest.

\subsection{Synthetic Experiments}

\begin{table}[t]
\centering
\begin{tabular}{c c l}
\hline
\hline
Order & Time Point &  Generative SBM Model \\
\hline
0 & 0 & $\bold{N_c = 4}, \; p_{in} = 0.25, \; p_{ex} = 0.05$ \\ 
1 & 16 & $\bold{N_c = 10}, \; p_{in} = 0.25, \; p_{ex} = 0.05$ \\
2 & 31 & $\bold{N_c = 2}, \; p_{in} = 0.5, \; p_{ex} = 0.05$ \\
3 & 61 & $\bold{N_c = 4}, \; p_{in} = 0.25, \; p_{ex} = 0.05$ \\
4 & 76 & $\bold{N_c = 10}, \; p_{in} = 0.25, \; p_{ex} = 0.05$ \\
5 & 91 & $\bold{N_c = 2}, \; p_{in} = 0.5, \; p_{ex} = 0.05$ \\      
6 & 106 & $\bold{N_c = 4}, \; p_{in} = 0.25, \; p_{ex} = 0.05$ \\
7 & 136 & $\bold{N_c = 10}, \; p_{in} = 0.25, \; p_{ex} = 0.05$ \\
\hline
\end{tabular}
\caption{Model changes in experiment \ref{exp:synth1} when only number of blocks ($N_c$) changes (Pure setting).
 }
\label{tab:SBM}
\vskip -0.3in
\end{table}

To demonstrate the performance of \method,  we design three controlled experiments. For these synthetic experiments, we use data generated from the Stochastic Block Model (SBM)~\cite{holland1983stochastic}. The number of communities $k$ as well as the number of nodes within each community can be specified through a size vector $\vec{s}\in \mathbb{R}^{k}$. In addition, the inter-community and intra-community connectivity for each block can be directly encoded in a symmetric probability matrix $P \in \mathbb{R}^{k \times k}$. For all experiments, we set the short term and the long term window to be 5 and 10 time steps respectively and use the entire Laplacian spectrum in \method. 

\begin{figure}[t]
    \begin{center}
        \centerline{\includegraphics[width=\columnwidth,trim={0in 0in 0 0 },clip]{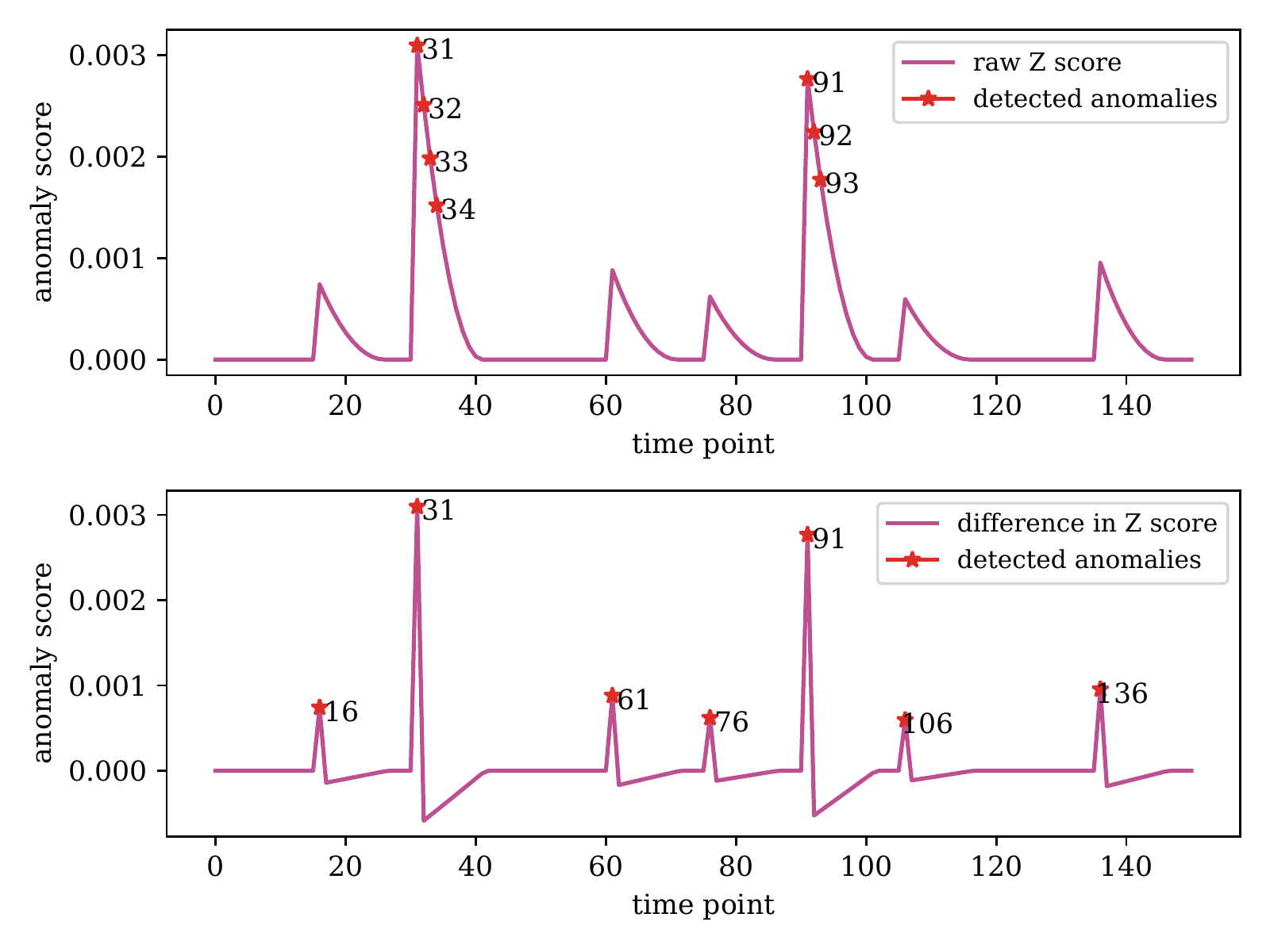}} \vskip -0.2in
        \caption{\method perfectly recovers the injected change points of Table~\ref{tab:SBM}. We visualize the predicted anomaly scores. Both the raw $Z$ score (top) and the difference in consecutive $Z$ scores, the $Z^*$ scores (bottom) are plotted.}
       \label{Fig:SBMscore}
    \end{center}
\end{figure}

\subsubsection{\textbf{Pure Setting}} 
\label{exp:synth1}
Here, we only introduce change points, where the adjustments in community structure persists until the next change point is reached. We generate a temporal network with 151 time points where each snapshot is produced through SBM parametrized by $\vec{s}$ and $P$. There are always 500 nodes per snapshot and the community change is described in Table~\ref{tab:SBM}. Here $N_{c}$ represents the number of equal sized communities in the snapshot, and $p_{in}, p_{ex}$ denotes the internal and external community connectivity probability respectively. We set the continuity rate~\cite{wang2017fast} to be 0 for change points and 1.0 elsewhere for the pure setting. 

The anomaly score predicted by LAD can be seen in Figure~\ref{Fig:SBMscore}. The top 7 most anomalous points correspond to the 7 ground truth points in Table~\ref{tab:SBM} while the other points have extremely low anomaly scores. This supports the empirical observation that the Laplacian spectrum is indeed sensitive to changes in community structure. EdgeMonitoring and \method both achieves perfect precision as both can reason with gradual changes over time. In contrast, both TENSORSPLAT and Activity vector can only recover some change points. As Activity vector uses the principal eigenvector of the adjacency matrix, it is unable to detect community changes as well as the Laplacian spectrum used in \method.

By examining Spearman rank correlations in Table~\ref{tab:rank}, we observe that \method predictions are most strongly correlated with the number of connected components while still having a positive correlations with other network properties such as transitivity. As \method captures high level graph structures, it is sensitive to the important properties in the dynamic graph of interest while not dependent on any single property.

Note that when using the raw $Z$ scores in Figure~\ref{Fig:SBMscore}, we observe declines in anomaly score after the change point. In a sliding window which contains graph snapshots from the previous graph generative process and the current one, the normal behavior vector is computed based on a mix of graphs from two generative processes thus leading to the observed declining anomalous scores after the change point. In comparison, using $Z^*$ scores makes \method  more robust under different choices of sliding windows. The negative $Z^*$ scores are only shown for illustrative purpose In Figure~\ref{Fig:SBMscore}, in other figures, they are set to 0. 

\begin{table}[t]
\centering
\begin{tabular}{l|c}
\hline
\hline
Graph Property & Spearman Rank Correlation \\
\hline 

\# of connected components &\textbf{ 17.0}\% \\
Transitivity & 11.8\% \\
\# of edges & 7.5\% \\
Average degree per node & 15.9\% \\
\hline
\end{tabular}
\caption{\method scores correlate well with the number of connected components when injected points are changing number of blocks in SBM (Table~\ref{tab:SBM}). Spearman rank correlation between \method and other graph properties is also reported.}
\label{tab:rank}
\vskip -0,3in
\end{table}

\begin{table}[t]
\centering
\begin{tabular}{c l l}
\hline
\hline
Time Point & Type &  Generative SBM Model \\
\hline
0 & start point & $N_c = 4, \; p_{in} = 0.25, \; p_{ex} = 0.05$ \\
16 & event & $N_c = 4, \; p_{in} = 0.25, \; \bold{p_{ex} = 0.15}$ \\
31 & change point & $\bold{N_c = 10}, \; p_{in} = 0.25, \; p_{ex} = 0.05$ \\
61 & event & $N_c = 10, \; p_{in} = 0.25, \; \bold{p_{ex} = 0.15}$ \\
76 & change point & $\bold{N_c = 2}, \; p_{in} = 0.5, \; p_{ex} = 0.05$ \\
91 & event & $N_c = 2, \; p_{in} = 0.5, \; \bold{p_{ex} = 0.15}$ \\
106 & change point & $\bold{N_c = 4}, \; p_{in} = 0.25, \; p_{ex} = 0.05$ \\
136 & event & $N_c = 4, \; p_{in} = 0.25, \;\bold{ p_{ex} = 0.15}$ \\
\hline
\end{tabular}
\caption{Model changes in experiment \ref{exp:synth12} where we have combination of event and change points (Hyprid setting). 
}
\label{tab:eventChange}
\vskip -0.3in
\end{table}

\subsubsection{\textbf{Hybrid Setting}}
\label{exp:synth12}

\begin{figure}[t]
    \begin{center}
        \centerline{\includegraphics[width=\columnwidth]{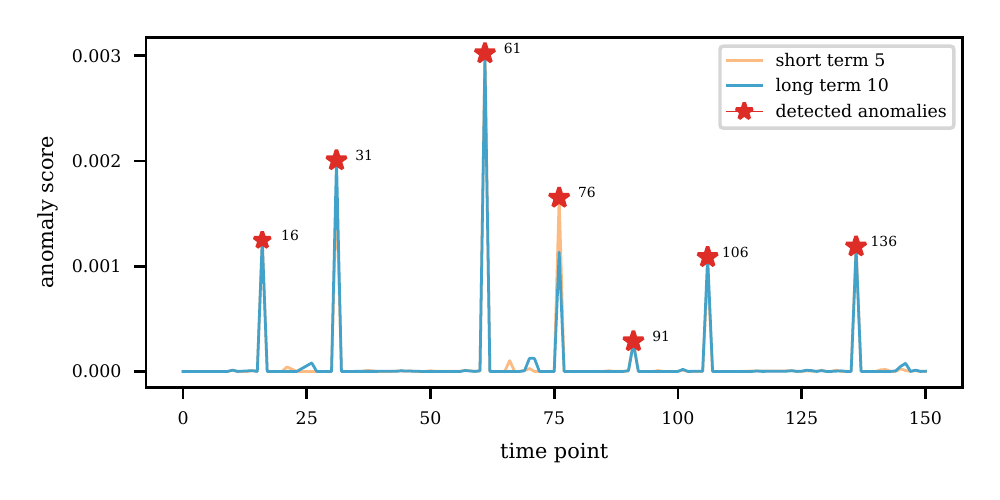}}    \vskip -0.3in
        \caption{\method perfectly recovers all \underline{events} and \underline{change} points defined in Table~\ref{tab:eventChange}. 
        }
        \label{Fig:eventChange}
    \end{center}
    \vskip -0.3in
\end{figure}

To study the effectiveness of \method in both change point and event detection, we generate a synthetic experiment with SBM where these two types of changes both exist. We generate events by strengthening the inter-community connectivity for that time point~(subsequent points are not affected). These events can correspond to sudden increased collaborations between usually separated communities such as political parties. The details regarding the generative process can be seen in Table~\ref{tab:eventChange}. We set the continuity rate~\cite{wang2017fast} to be 0 for change points and 0.9 elsewhere for the hybrid setting. From Figure~\ref{Fig:eventChange}, we observe that \method is able to perfectly recover all events and change points. 

{\footnotesize
    \begin{table}[t]
        \centering
        \begin{tabular}{c | c c c c c}
        \hline
        \hline
        Metric & \multicolumn{3}{c}{Hits @ 7} & Hits @ 10 & Hits @ 2\\
        \hline
        Dataset & Pure & Hybrid & Resampled & UCI & Senate \\
        \hline
        \hline
        \rowcolor{gray!20}LAD~(ours) & \textbf{100\%} & \textbf{100\%} & \textbf{100\%} & \textbf{50\%} & \textbf{100\%} \\
        EdgeMonitoring~\cite{wang2017fast} & \textbf{100\%} & \textbf{100\%} & 0\% & 0\% & \textbf{100\%} \\
        Activity vector~\cite{ide2004eigenspace} & 71.4\% & 0\% & 0\% & \textbf{50\%} & 50\% \\
        TENSORSPLAT~\cite{koutra2012tensorsplat} & 28.5\%& 14.2\% & 57.1\% & 0\% & 50\% \\
        \hline
        \end{tabular}
        \caption{\method  consistently finds significant anomalies across different datasets and outperforms alternative approaches. The hybrid experiment refers to the event and change point detection experiment.}
        \label{tab:results}
        \vskip -0.3in
    \end{table}
}

\subsubsection{\textbf{Resampled Setting}}
In this setting, we use a constant continuity rate of 0 for all time steps, i.e. the graph is resampled from the generative process at each step. In real world graphs, majority of the edges might not persist over consecutive time steps but rather determined by the underlying community structure such as a club or class. 
For the generation parameters and change points' details, we use an identical setup as the Hybrid Setting. The complete resampling from the SBM model can be considered as a node level permutation within each community. As discussed in Section~\ref{Sec:method}, the eigenvalues of the Laplacian is node permutation invariant. Indeed, \method outperforms all baselines. In comparison, EdgeMonitoring selects specific pairs of node pairs to track over time thus it is susceptible to node permutation and resulted in poor performance.

\subsection{Real-world Experiments}
Here, we evaluate the performance of \method on two real-world benchmark datasets, as well as an original dataset. we report the performance of \method and all baselines in Table~\ref{tab:results}. For UCI Message and Senate co-sponsorship experiments, \method is able to achieve strong performance using only the top 6 eigenvalues. For Canadian bill voting network, we report the LAD performance with the top 338 eigenvalues~\footnote{338 is the number of seats in the House of Commons of Canada}. 

\subsubsection{\textbf{UCI Message}} 
The UCI Message dataset is a directed and weighted network based on an online community of students at the University of California, Irvine. Each node represents a user and each edge encodes a message interaction from one user to another. The weight of each edge represents the number of characters sent in the message. When an user account is created, a self edge with unit weight is added. A total of 1,899 users was recorded. The network data covers the period from April to October 2004 and spans 196 days. 
In this work, we treat each day as an individual time point. There are 59,835 total messages sent across all time stamps and 20,296 unique messages. To ensure privacy protection, all individual identifiers such as usernames, email and IP addresses were removed thus we use the dataset description provided in~\cite{panzarasa2009patterns} to find significant events in the dataset. We select the short term window to be 7 days as suggested by Panzarasa et al. The long term window is then selected to be 14 days or two weeks. 

\begin{figure}[t]
    \begin{center}
        \centerline{\includegraphics[width=\columnwidth]{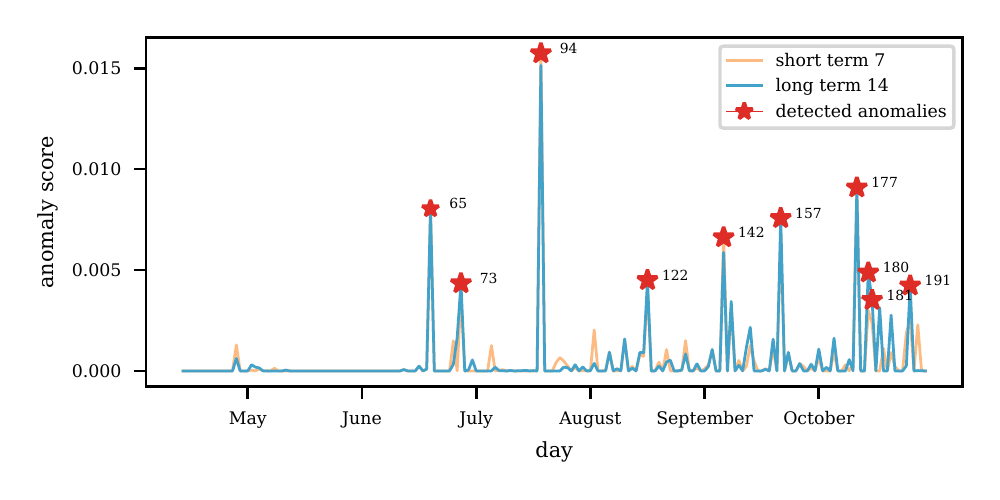}} \vskip -0.2in
        \caption{\method correctly detects the end of the university spring term and one day before the start of the fall term in the UCI message dataset.
        }
       \label{Fig:UCIscore}
    \end{center}
    \vskip -0.3in
\end{figure}

Figure~\ref{Fig:UCIscore} shows the $Z^*$ scores predicted from both the short term and long term window~(with top 6 eigenvalues). Panzarasa et al. mentioned that day 65\footnote{we set the index to start at 0} is the end of spring term and day 158 is the start of the fall term. Day 65 is correctly predicted by \method and activity vector while the top 10 predictions from EdgeMonitoring and TENSORSPLAT do not include either days. However, \method predicts day 157 as anomalous which corresponds to the day prior to the start of the fall term. It is likely that anomalous message behaviors are exhibited before school starts. As the edge weight~(number of characters in messages) shows important connections between users in social networks, the $Z$ scores has strongest correlation to the average edge weight per snapshot in this dataset.

\subsubsection{\textbf{Senate co-sponsorship Network}}

\begin{figure}[t]
    \begin{center}
        \centerline{\includegraphics[width=\columnwidth]{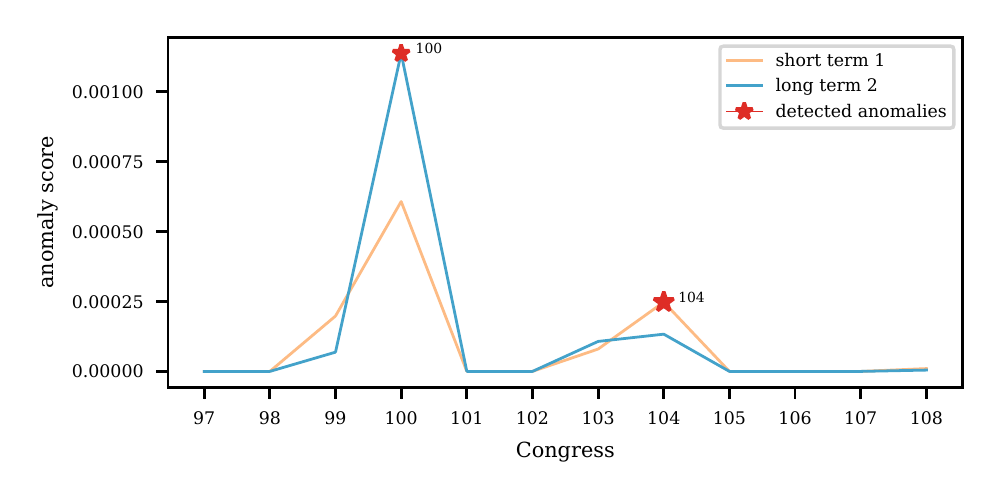}} \vskip -0.2in
        \caption{\method correctly detects the 100th and 104th congress as the top 2 most anomalous points.}
       \label{Fig:senateScore}
    \end{center}
    \vskip -0.3in
\end{figure}

Senate co-sponsorship network~\cite{fowler2006legislative} examines social connections between legislators from their co-sponsorship relations on bills during the 93rd-108th Congress. An edge is formed between two congresspersons if they are cosponsors on a bill. Bills are grouped into temporal snapshots biannually~(time frame for each graph) and co-sponsors on a bill form a clique. Similar to~\cite{wang2017fast}, we start from the 97th Congress as full amendments data is available from there onward. Fowler~\cite{fowler2006legislative} pointed out that the 104th Congress corresponds to "a Republican Revolution" which "caused a dramatic change in the partisan and seniority compositions." Wang et al. also stated that the 104th Congress has the lowest clustering coefficient thus can be seen as a low point in collaboration while the 100th Congress has the highest clustering coefficient which signals significant collaboration. 

From Figure~\ref{Fig:senateScore}, it is clear that \method can easily detect the above change points~(using only the top 6 singular values). Variations of \method that only utilizes the short term or the long term window are able to identify both points too. Wang et al. mentioned that EdgeMonitoring~\cite{wang2017fast} and LetoChange \cite{peel2015detecting} are able to detect both change points while DeltaCon~\cite{koutra2016deltacon} only predicts the 104th Congress.

The activity vector method requires full SVD which are computationally expensive and it is only able to detect the 100th Congress. The anomaly scores output by the activity vector method have similar magnitude thus making it difficult to identify change points. However, if we augment the activity vector method to use \method pipeline and aggregates two sliding windows, it is able to correctly predict both change points. This shows that aggregating the output from two sliding windows can improve empirical performance with a different graph embedding technique.

\subsubsection{\textbf{Canadian bill voting network}}

To understand the temporal change in Canadian Parliament environment, we extracted open information\footnote{extracted from \url{https://openparliament.ca/}} to form the Canadian Parliament bill voting network. The Canadian Parliament consists of 338 Members of Parliament~(MPs), each representing an electoral district, who are elected for four years and can be re-elected~\cite{Parliament}. In the included time frame from 2006 to 2019, the increase in the number of electoral seats resulted in parliaments with different amounts of MPs; since 2015 the House of Commons has grown from 308 seats to 338. Naturally, the number of nodes in the network fluctuates from year to year. 2 year and 4 years are used as duration for the short term and long term windows respectively. We consider the MPs that voted yes for a bill to have a positive relation with the sponsor. In this way, a directed edge is then formed from a voter MP $u$ to the sponsor MP $v$ in a given graph snapshot. Each edge is then weighted by the number of times that $u$ voted positively for $v$.

\begin{figure}[t]
    \begin{center}        \centerline{\includegraphics[width=\columnwidth]{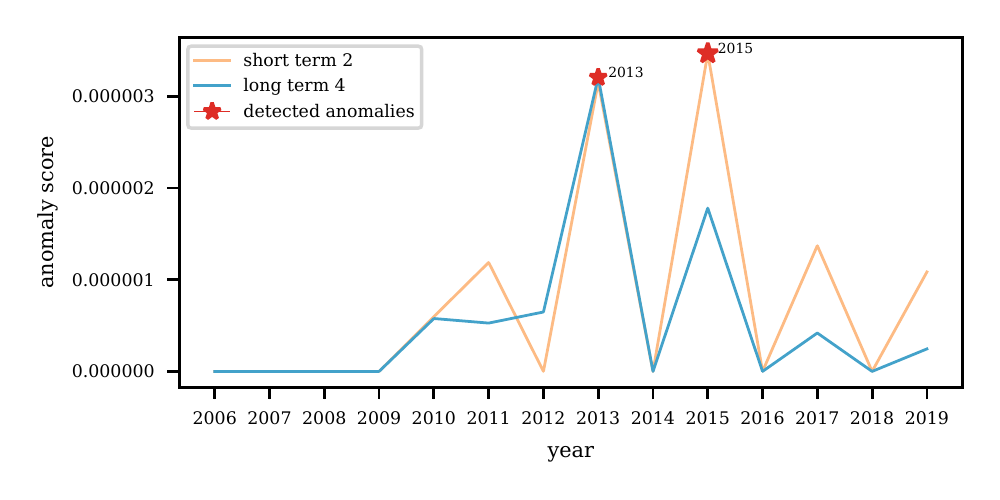}} \vskip -0.2in
        \caption{\method  predicts year 2013 and 2015 as anomalous years for the Canadian bill voting network.
        }
      \label{Fig:canVotescore}
    \end{center}
    \vspace{-20pt}
\end{figure}

\method detects 2013 and 2015 as two significant change points. Figure~\ref{Fig:canVotescore} shows the $Z^*$ scores predicted by \method. 2013 is often considered as the year where cross party cohesion against Conservatives begins to decline. Prior to 2013, the Liberal and New Democratic Parties had more unity in voting patterns. As a new party leader~(Justin Trudeau) is elected in the Liberal party in 2013, changes in voting patterns are observed~\cite{marland2013political}. More details regarding data mining are discussed in Appendix~\ref{app:canadian}. In 2015, the house of common increased the number of constituencies from 308 to 338. Equally, on October 19th 2015 an election took place and the Liberal party won an additional 148 seats, with a total of 184 seats forming a majority government led by Justin Trudeau. Prior to 2015, the Liberal party was divided and however during this election things changed and literature shows the unified campaign at a local level from different members of parliament across the country \cite{cross2016importance}.





\section{Discussion}

Table~\ref{tab:results} summarizes the empirical performances of \method and its comparison to alternative methods. We observe that \method has the best performance across all datasets. In this section, we discuss several experimental observations and provide  intuitions on the experimental results.  


In the UCI message and senate co-sponsorship experiments, \method achieves strong performance using only the top 6 singular values. It demonstrates that in real world datasets, a low rank truncated SVD is often sufficient to capture rich graph information. Together with efficient SVD computation techniques, \method can be scaled to large networks. In practice, the rank of the truncated SVD can be determined by the available computational resources. 

By using Spearman rank correlation, we observe that \method is not dependent on any one particular graph statistics. Empirically, we observe that \method correlates with different graph properties depending on the network of interest. This coincide with our intuition that any type of significant change in the graph structure could disrupt the singular values. These experimental results suggest that \method can capture changes of different nature in the graph structure as it tracks the compression loss of the Laplacian matrix to low rank approximations. 


Lastly, synthetic experiment results suggest that \method can be used in a hybrid environment where both change points or events can occur. Therefore, \method is suitable to detect anomalous time points where it is possible for both one time events or fundamental changes in the network generative process to occur. It is often difficult to know beforehand what types of anomaly would appear in a dynamic network, \method would provide an effective and efficient solution with no assumptions on the anomaly type.

\section{Conclusion}
We proposed a novel spectral based method capable of both change point and event detection, called \method. The core idea of \method is to summarize entire graph snapshots into low dimensional embeddings through the singular values of the graph Laplacian. Different than previous approaches, \method explicitly models the short term and the long term behavior of the dynamic graph and aggregates both perspectives. When compared to existing methods, \method outperforms all alternative methods on synthetic experiments and \method finds well-known events in three real networks.

\begin{acks}
We thank Wang et al.~\cite{wang2017fast} for providing the implementation of EdgeMonitoring. We thank Professor Jean-François Godbout at University of Montreal for insightful discussions about anomalous voting patterns in the Canadian Parliament voting network. We thank Kuan-Chieh (Jackson) Wang for helpful feedback and discussions.  This research is supported by the Canadian Institute for Advanced Research (CIFAR AI chair program).
\end{acks}

\bibliographystyle{ACM-Reference-Format}
\bibliography{main}


\begin{thebibliography}{51}


\ifx \showCODEN    \undefined \def \showCODEN     #1{\unskip}     \fi
\ifx \showDOI      \undefined \def \showDOI       #1{#1}\fi
\ifx \showISBNx    \undefined \def \showISBNx     #1{\unskip}     \fi
\ifx \showISBNxiii \undefined \def \showISBNxiii  #1{\unskip}     \fi
\ifx \showISSN     \undefined \def \showISSN      #1{\unskip}     \fi
\ifx \showLCCN     \undefined \def \showLCCN      #1{\unskip}     \fi
\ifx \shownote     \undefined \def \shownote      #1{#1}          \fi
\ifx \showarticletitle \undefined \def \showarticletitle #1{#1}   \fi
\ifx \showURL      \undefined \def \showURL       {\relax}        \fi
\providecommand\bibfield[2]{#2}
\providecommand\bibinfo[2]{#2}
\providecommand\natexlab[1]{#1}
\providecommand\showeprint[2][]{arXiv:#2}

\bibitem[\protect\citeauthoryear{Aharon, Elad, and Bruckstein}{Aharon
  et~al\mbox{.}}{2006}]%
        {aharon2006k}
\bibfield{author}{\bibinfo{person}{Michal Aharon}, \bibinfo{person}{Michael
  Elad}, {and} \bibinfo{person}{Alfred Bruckstein}.}
  \bibinfo{year}{2006}\natexlab{}.
\newblock \showarticletitle{K-SVD: An algorithm for designing overcomplete
  dictionaries for sparse representation}.
\newblock \bibinfo{journal}{\emph{IEEE Transactions on signal processing}}
  \bibinfo{volume}{54}, \bibinfo{number}{11} (\bibinfo{year}{2006}),
  \bibinfo{pages}{4311--4322}.
\newblock


\bibitem[\protect\citeauthoryear{Akoglu and Faloutsos}{Akoglu and
  Faloutsos}{2010}]%
        {akoglu2010event}
\bibfield{author}{\bibinfo{person}{Leman Akoglu} {and}
  \bibinfo{person}{Christos Faloutsos}.} \bibinfo{year}{2010}\natexlab{}.
\newblock \showarticletitle{Event detection in time series of mobile
  communication graphs}. In \bibinfo{booktitle}{\emph{Army science
  conference}}, Vol.~\bibinfo{volume}{1}.
\newblock


\bibitem[\protect\citeauthoryear{Benjamin~Erichson, Brunton, and
  Nathan~Kutz}{Benjamin~Erichson et~al\mbox{.}}{2017}]%
        {benjamin2017compressed}
\bibfield{author}{\bibinfo{person}{N Benjamin~Erichson},
  \bibinfo{person}{Steven~L Brunton}, {and} \bibinfo{person}{J Nathan~Kutz}.}
  \bibinfo{year}{2017}\natexlab{}.
\newblock \showarticletitle{Compressed singular value decomposition for image
  and video processing}. In \bibinfo{booktitle}{\emph{Proceedings of the IEEE
  International Conference on Computer Vision Workshops}}.
  \bibinfo{pages}{1880--1888}.
\newblock


\bibitem[\protect\citeauthoryear{Berry}{Berry}{1992}]%
        {berry1992large}
\bibfield{author}{\bibinfo{person}{Michael~W Berry}.}
  \bibinfo{year}{1992}\natexlab{}.
\newblock \showarticletitle{Large-scale sparse singular value computations}.
\newblock \bibinfo{journal}{\emph{The International Journal of Supercomputing
  Applications}} \bibinfo{volume}{6}, \bibinfo{number}{1}
  (\bibinfo{year}{1992}), \bibinfo{pages}{13--49}.
\newblock


\bibitem[\protect\citeauthoryear{Bonacich}{Bonacich}{1987}]%
        {bonacich1987power}
\bibfield{author}{\bibinfo{person}{Phillip Bonacich}.}
  \bibinfo{year}{1987}\natexlab{}.
\newblock \showarticletitle{Power and centrality: A family of measures}.
\newblock \bibinfo{journal}{\emph{American journal of sociology}}
  \bibinfo{volume}{92}, \bibinfo{number}{5} (\bibinfo{year}{1987}),
  \bibinfo{pages}{1170--1182}.
\newblock


\bibitem[\protect\citeauthoryear{Breunig, Kriegel, Ng, and Sander}{Breunig
  et~al\mbox{.}}{2000}]%
        {breunig2000lof}
\bibfield{author}{\bibinfo{person}{Markus~M Breunig},
  \bibinfo{person}{Hans-Peter Kriegel}, \bibinfo{person}{Raymond~T Ng}, {and}
  \bibinfo{person}{J{\"o}rg Sander}.} \bibinfo{year}{2000}\natexlab{}.
\newblock \showarticletitle{LOF: identifying density-based local outliers}. In
  \bibinfo{booktitle}{\emph{Proceedings of the 2000 ACM SIGMOD international
  conference on Management of data}}. \bibinfo{pages}{93--104}.
\newblock


\bibitem[\protect\citeauthoryear{Bro}{Bro}{1997}]%
        {bro1997parafac}
\bibfield{author}{\bibinfo{person}{Rasmus Bro}.}
  \bibinfo{year}{1997}\natexlab{}.
\newblock \showarticletitle{PARAFAC. Tutorial and applications}.
\newblock \bibinfo{journal}{\emph{Chemometrics and intelligent laboratory
  systems}} \bibinfo{volume}{38}, \bibinfo{number}{2} (\bibinfo{year}{1997}),
  \bibinfo{pages}{149--171}.
\newblock


\bibitem[\protect\citeauthoryear{Broido and Clauset}{Broido and
  Clauset}{2019}]%
        {broido2019scale}
\bibfield{author}{\bibinfo{person}{Anna~D Broido} {and} \bibinfo{person}{Aaron
  Clauset}.} \bibinfo{year}{2019}\natexlab{}.
\newblock \showarticletitle{Scale-free networks are rare}.
\newblock \bibinfo{journal}{\emph{Nature communications}} \bibinfo{volume}{10},
  \bibinfo{number}{1} (\bibinfo{year}{2019}), \bibinfo{pages}{1017}.
\newblock


\bibitem[\protect\citeauthoryear{Chen, Hendrix, and Samatova}{Chen
  et~al\mbox{.}}{2012}]%
        {chen2012community}
\bibfield{author}{\bibinfo{person}{Zhengzhang Chen}, \bibinfo{person}{William
  Hendrix}, {and} \bibinfo{person}{Nagiza~F Samatova}.}
  \bibinfo{year}{2012}\natexlab{}.
\newblock \showarticletitle{Community-based anomaly detection in evolutionary
  networks}.
\newblock \bibinfo{journal}{\emph{Journal of Intelligent Information Systems}}
  \bibinfo{volume}{39}, \bibinfo{number}{1} (\bibinfo{year}{2012}),
  \bibinfo{pages}{59--85}.
\newblock


\bibitem[\protect\citeauthoryear{Chung and Graham}{Chung and Graham}{1997}]%
        {chung1997spectral}
\bibfield{author}{\bibinfo{person}{Fan~RK Chung} {and}
  \bibinfo{person}{Fan~Chung Graham}.} \bibinfo{year}{1997}\natexlab{}.
\newblock \bibinfo{booktitle}{\emph{Spectral graph theory}}.
\newblock Number~92. \bibinfo{publisher}{American Mathematical Soc.}
\newblock


\bibitem[\protect\citeauthoryear{Commons}{Commons}{[n.d.]}]%
        {Parliament}
\bibfield{author}{\bibinfo{person}{Our Commons}.}
  \bibinfo{year}{[n.d.]}\natexlab{}.
\newblock \bibinfo{booktitle}{\emph{In the House}}.
\newblock
\urldef\tempurl%
\url{https://www.ourcommons.ca/en}
\showURL{%
Retrieved February 9, 2020 from \tempurl}


\bibitem[\protect\citeauthoryear{Conley}{Conley}{2011}]%
        {conley2011legislative}
\bibfield{author}{\bibinfo{person}{Richard~S Conley}.}
  \bibinfo{year}{2011}\natexlab{}.
\newblock \showarticletitle{Legislative Activity in the Canadian House of
  Commons: Does Majority or Minority Government Matter?}
\newblock \bibinfo{journal}{\emph{American Review of Canadian Studies}}
  \bibinfo{volume}{41}, \bibinfo{number}{4} (\bibinfo{year}{2011}),
  \bibinfo{pages}{422--437}.
\newblock


\bibitem[\protect\citeauthoryear{Cross}{Cross}{2016}]%
        {cross2016importance}
\bibfield{author}{\bibinfo{person}{William Cross}.}
  \bibinfo{year}{2016}\natexlab{}.
\newblock \showarticletitle{The Importance of Local Party Activity in
  Understanding Canadian Politics: Winning from the Ground Up in the 2015
  Federal Election: Presidential Address to the Canadian Political Science
  Association Calgary, 31 May 2016}.
\newblock \bibinfo{journal}{\emph{Canadian Journal of Political Science/Revue
  canadienne de science politique}} \bibinfo{volume}{49}, \bibinfo{number}{4}
  (\bibinfo{year}{2016}), \bibinfo{pages}{601--620}.
\newblock


\bibitem[\protect\citeauthoryear{Eswaran, Faloutsos, Guha, and Mishra}{Eswaran
  et~al\mbox{.}}{2018}]%
        {eswaran2018spotlight}
\bibfield{author}{\bibinfo{person}{Dhivya Eswaran}, \bibinfo{person}{Christos
  Faloutsos}, \bibinfo{person}{Sudipto Guha}, {and} \bibinfo{person}{Nina
  Mishra}.} \bibinfo{year}{2018}\natexlab{}.
\newblock \showarticletitle{Spotlight: Detecting anomalies in streaming
  graphs}. In \bibinfo{booktitle}{\emph{Proceedings of the 24th ACM SIGKDD
  International Conference on Knowledge Discovery \& Data Mining}}.
  \bibinfo{pages}{1378--1386}.
\newblock


\bibitem[\protect\citeauthoryear{Fowler}{Fowler}{2006}]%
        {fowler2006legislative}
\bibfield{author}{\bibinfo{person}{James~H Fowler}.}
  \bibinfo{year}{2006}\natexlab{}.
\newblock \showarticletitle{Legislative cosponsorship networks in the US House
  and Senate}.
\newblock \bibinfo{journal}{\emph{Social Networks}} \bibinfo{volume}{28},
  \bibinfo{number}{4} (\bibinfo{year}{2006}), \bibinfo{pages}{454--465}.
\newblock


\bibitem[\protect\citeauthoryear{Gahrooei and Paynabar}{Gahrooei and
  Paynabar}{2018}]%
        {gahrooei2018change}
\bibfield{author}{\bibinfo{person}{Mostafa~Reisi Gahrooei} {and}
  \bibinfo{person}{Kamran Paynabar}.} \bibinfo{year}{2018}\natexlab{}.
\newblock \showarticletitle{Change detection in a dynamic stream of attributed
  networks}.
\newblock \bibinfo{journal}{\emph{Journal of Quality Technology}}
  \bibinfo{volume}{50}, \bibinfo{number}{4} (\bibinfo{year}{2018}),
  \bibinfo{pages}{418--430}.
\newblock


\bibitem[\protect\citeauthoryear{Garner and Letki}{Garner and Letki}{2005}]%
        {garner2005party}
\bibfield{author}{\bibinfo{person}{Christopher Garner} {and}
  \bibinfo{person}{Natalia Letki}.} \bibinfo{year}{2005}\natexlab{}.
\newblock \showarticletitle{Party structure and backbench dissent in the
  Canadian and British Parliaments}.
\newblock \bibinfo{journal}{\emph{Canadian Journal of Political Science/Revue
  canadienne de science politique}} \bibinfo{volume}{38}, \bibinfo{number}{2}
  (\bibinfo{year}{2005}), \bibinfo{pages}{463--482}.
\newblock


\bibitem[\protect\citeauthoryear{Goh, Kahng, and Kim}{Goh
  et~al\mbox{.}}{2001}]%
        {goh2001universal}
\bibfield{author}{\bibinfo{person}{K-I Goh}, \bibinfo{person}{Byungnam Kahng},
  {and} \bibinfo{person}{Doochul Kim}.} \bibinfo{year}{2001}\natexlab{}.
\newblock \showarticletitle{Universal behavior of load distribution in
  scale-free networks}.
\newblock \bibinfo{journal}{\emph{Physical review letters}}
  \bibinfo{volume}{87}, \bibinfo{number}{27} (\bibinfo{year}{2001}),
  \bibinfo{pages}{278701}.
\newblock


\bibitem[\protect\citeauthoryear{Golub and Reinsch}{Golub and Reinsch}{1971}]%
        {golub1971singular}
\bibfield{author}{\bibinfo{person}{Gene~H Golub} {and}
  \bibinfo{person}{Christian Reinsch}.} \bibinfo{year}{1971}\natexlab{}.
\newblock \showarticletitle{Singular value decomposition and least squares
  solutions}.
\newblock In \bibinfo{booktitle}{\emph{Linear Algebra}}.
  \bibinfo{publisher}{Springer}, \bibinfo{pages}{134--151}.
\newblock


\bibitem[\protect\citeauthoryear{Hagen and Kahng}{Hagen and Kahng}{1992}]%
        {hagen1992new}
\bibfield{author}{\bibinfo{person}{Lars Hagen} {and} \bibinfo{person}{Andrew~B
  Kahng}.} \bibinfo{year}{1992}\natexlab{}.
\newblock \showarticletitle{New spectral methods for ratio cut partitioning and
  clustering}.
\newblock \bibinfo{journal}{\emph{IEEE transactions on computer-aided design of
  integrated circuits and systems}} \bibinfo{volume}{11}, \bibinfo{number}{9}
  (\bibinfo{year}{1992}), \bibinfo{pages}{1074--1085}.
\newblock


\bibitem[\protect\citeauthoryear{Halko, Martinsson, and Tropp}{Halko
  et~al\mbox{.}}{2011}]%
        {halko2011finding}
\bibfield{author}{\bibinfo{person}{Nathan Halko}, \bibinfo{person}{Per-Gunnar
  Martinsson}, {and} \bibinfo{person}{Joel~A Tropp}.}
  \bibinfo{year}{2011}\natexlab{}.
\newblock \showarticletitle{Finding structure with randomness: Probabilistic
  algorithms for constructing approximate matrix decompositions}.
\newblock \bibinfo{journal}{\emph{SIAM review}} \bibinfo{volume}{53},
  \bibinfo{number}{2} (\bibinfo{year}{2011}), \bibinfo{pages}{217--288}.
\newblock


\bibitem[\protect\citeauthoryear{Harshman et~al\mbox{.}}{Harshman
  et~al\mbox{.}}{1970}]%
        {harshman1970foundations}
\bibfield{author}{\bibinfo{person}{Richard~A Harshman} {et~al\mbox{.}}}
  \bibinfo{year}{1970}\natexlab{}.
\newblock \showarticletitle{Foundations of the PARAFAC procedure: Models and
  conditions for an" explanatory" multimodal factor analysis}.
\newblock  (\bibinfo{year}{1970}).
\newblock


\bibitem[\protect\citeauthoryear{Holland, Laskey, and Leinhardt}{Holland
  et~al\mbox{.}}{1983}]%
        {holland1983stochastic}
\bibfield{author}{\bibinfo{person}{Paul~W Holland},
  \bibinfo{person}{Kathryn~Blackmond Laskey}, {and} \bibinfo{person}{Samuel
  Leinhardt}.} \bibinfo{year}{1983}\natexlab{}.
\newblock \showarticletitle{Stochastic blockmodels: First steps}.
\newblock \bibinfo{journal}{\emph{Social networks}} \bibinfo{volume}{5},
  \bibinfo{number}{2} (\bibinfo{year}{1983}), \bibinfo{pages}{109--137}.
\newblock


\bibitem[\protect\citeauthoryear{Id{\'e} and Kashima}{Id{\'e} and
  Kashima}{2004}]%
        {ide2004eigenspace}
\bibfield{author}{\bibinfo{person}{Tsuyoshi Id{\'e}} {and}
  \bibinfo{person}{Hisashi Kashima}.} \bibinfo{year}{2004}\natexlab{}.
\newblock \showarticletitle{Eigenspace-based anomaly detection in computer
  systems}. In \bibinfo{booktitle}{\emph{Proceedings of the tenth ACM SIGKDD
  international conference on Knowledge discovery and data mining}}. ACM,
  \bibinfo{pages}{440--449}.
\newblock


\bibitem[\protect\citeauthoryear{Kolda and Bader}{Kolda and Bader}{2009}]%
        {kolda2009tensor}
\bibfield{author}{\bibinfo{person}{Tamara~G Kolda} {and}
  \bibinfo{person}{Brett~W Bader}.} \bibinfo{year}{2009}\natexlab{}.
\newblock \showarticletitle{Tensor decompositions and applications}.
\newblock \bibinfo{journal}{\emph{SIAM review}} \bibinfo{volume}{51},
  \bibinfo{number}{3} (\bibinfo{year}{2009}), \bibinfo{pages}{455--500}.
\newblock


\bibitem[\protect\citeauthoryear{Kossaifi, Panagakis, Anandkumar, and
  Pantic}{Kossaifi et~al\mbox{.}}{2019}]%
        {kossaifi2019tensorly}
\bibfield{author}{\bibinfo{person}{Jean Kossaifi}, \bibinfo{person}{Yannis
  Panagakis}, \bibinfo{person}{Anima Anandkumar}, {and} \bibinfo{person}{Maja
  Pantic}.} \bibinfo{year}{2019}\natexlab{}.
\newblock \showarticletitle{Tensorly: Tensor learning in python}.
\newblock \bibinfo{journal}{\emph{The Journal of Machine Learning Research}}
  \bibinfo{volume}{20}, \bibinfo{number}{1} (\bibinfo{year}{2019}),
  \bibinfo{pages}{925--930}.
\newblock


\bibitem[\protect\citeauthoryear{Koutra, Papalexakis, and Faloutsos}{Koutra
  et~al\mbox{.}}{2012}]%
        {koutra2012tensorsplat}
\bibfield{author}{\bibinfo{person}{Danai Koutra}, \bibinfo{person}{Evangelos~E
  Papalexakis}, {and} \bibinfo{person}{Christos Faloutsos}.}
  \bibinfo{year}{2012}\natexlab{}.
\newblock \showarticletitle{Tensorsplat: Spotting latent anomalies in time}. In
  \bibinfo{booktitle}{\emph{2012 16th Panhellenic Conference on Informatics}}.
  IEEE, \bibinfo{pages}{144--149}.
\newblock


\bibitem[\protect\citeauthoryear{Koutra, Shah, Vogelstein, Gallagher, and
  Faloutsos}{Koutra et~al\mbox{.}}{2016}]%
        {koutra2016deltacon}
\bibfield{author}{\bibinfo{person}{Danai Koutra}, \bibinfo{person}{Neil Shah},
  \bibinfo{person}{Joshua~T Vogelstein}, \bibinfo{person}{Brian Gallagher},
  {and} \bibinfo{person}{Christos Faloutsos}.} \bibinfo{year}{2016}\natexlab{}.
\newblock \showarticletitle{Deltacon: Principled massive-graph similarity
  function with attribution}.
\newblock \bibinfo{journal}{\emph{ACM Transactions on Knowledge Discovery from
  Data (TKDD)}} \bibinfo{volume}{10}, \bibinfo{number}{3}
  (\bibinfo{year}{2016}), \bibinfo{pages}{1--43}.
\newblock


\bibitem[\protect\citeauthoryear{Li, Li, Han, and Lee}{Li
  et~al\mbox{.}}{2009}]%
        {li2009temporal}
\bibfield{author}{\bibinfo{person}{Xiaolei Li}, \bibinfo{person}{Zhenhui Li},
  \bibinfo{person}{Jiawei Han}, {and} \bibinfo{person}{Jae-Gil Lee}.}
  \bibinfo{year}{2009}\natexlab{}.
\newblock \showarticletitle{Temporal outlier detection in vehicle traffic
  data}. In \bibinfo{booktitle}{\emph{2009 IEEE 25th International Conference
  on Data Engineering}}. IEEE, \bibinfo{pages}{1319--1322}.
\newblock


\bibitem[\protect\citeauthoryear{Macfarlane}{Macfarlane}{2019}]%
        {macfarlane2019renewed}
\bibfield{author}{\bibinfo{person}{Emmett Macfarlane}.}
  \bibinfo{year}{2019}\natexlab{}.
\newblock \bibinfo{booktitle}{\emph{The Renewed Canadian Senate: Organizational
  Challenges and Relations with the Government}}.
\newblock \bibinfo{publisher}{Institute for Research on Public Policy}.
\newblock


\bibitem[\protect\citeauthoryear{Marland}{Marland}{2013}]%
        {marland2013political}
\bibfield{author}{\bibinfo{person}{Alex Marland}.}
  \bibinfo{year}{2013}\natexlab{}.
\newblock \showarticletitle{What is a political brand?: Justin Trudeau and the
  theory of political branding}. In \bibinfo{booktitle}{\emph{annual meeting of
  the Canadian Communication Association and the Canadian Political Science
  Association, University of Victoria, British Columbia, June}},
  Vol.~\bibinfo{volume}{6}.
\newblock


\bibitem[\protect\citeauthoryear{Miller, Arcolano, Beard, Kepner, Schmidt,
  Bliss, and Wolfe}{Miller et~al\mbox{.}}{2012}]%
        {miller2012scalable}
\bibfield{author}{\bibinfo{person}{Benjamin~A Miller},
  \bibinfo{person}{Nicholas Arcolano}, \bibinfo{person}{Michelle~S Beard},
  \bibinfo{person}{Jeremy Kepner}, \bibinfo{person}{Matthew~C Schmidt},
  \bibinfo{person}{Nadya~T Bliss}, {and} \bibinfo{person}{Patrick~J Wolfe}.}
  \bibinfo{year}{2012}\natexlab{}.
\newblock \showarticletitle{A scalable signal processing architecture for
  massive graph analysis}. In \bibinfo{booktitle}{\emph{2012 IEEE International
  Conference on Acoustics, Speech and Signal Processing (ICASSP)}}. IEEE,
  \bibinfo{pages}{5329--5332}.
\newblock


\bibitem[\protect\citeauthoryear{Page, Brin, Motwani, and Winograd}{Page
  et~al\mbox{.}}{1999}]%
        {page1999pagerank}
\bibfield{author}{\bibinfo{person}{Lawrence Page}, \bibinfo{person}{Sergey
  Brin}, \bibinfo{person}{Rajeev Motwani}, {and} \bibinfo{person}{Terry
  Winograd}.} \bibinfo{year}{1999}\natexlab{}.
\newblock \bibinfo{booktitle}{\emph{The pagerank citation ranking: Bringing
  order to the web.}}
\newblock \bibinfo{type}{{T}echnical {R}eport}. \bibinfo{institution}{Stanford
  InfoLab}.
\newblock


\bibitem[\protect\citeauthoryear{Panzarasa, Opsahl, and Carley}{Panzarasa
  et~al\mbox{.}}{2009}]%
        {panzarasa2009patterns}
\bibfield{author}{\bibinfo{person}{Pietro Panzarasa}, \bibinfo{person}{Tore
  Opsahl}, {and} \bibinfo{person}{Kathleen~M Carley}.}
  \bibinfo{year}{2009}\natexlab{}.
\newblock \showarticletitle{Patterns and dynamics of users' behavior and
  interaction: Network analysis of an online community}.
\newblock \bibinfo{journal}{\emph{Journal of the American Society for
  Information Science and Technology}} \bibinfo{volume}{60},
  \bibinfo{number}{5} (\bibinfo{year}{2009}), \bibinfo{pages}{911--932}.
\newblock


\bibitem[\protect\citeauthoryear{Pedregosa, Varoquaux, Gramfort, Michel,
  Thirion, Grisel, Blondel, Prettenhofer, Weiss, Dubourg, Vanderplas, Passos,
  Cournapeau, Brucher, Perrot, and Duchesnay}{Pedregosa et~al\mbox{.}}{2011}]%
        {scikit-learn}
\bibfield{author}{\bibinfo{person}{F. Pedregosa}, \bibinfo{person}{G.
  Varoquaux}, \bibinfo{person}{A. Gramfort}, \bibinfo{person}{V. Michel},
  \bibinfo{person}{B. Thirion}, \bibinfo{person}{O. Grisel},
  \bibinfo{person}{M. Blondel}, \bibinfo{person}{P. Prettenhofer},
  \bibinfo{person}{R. Weiss}, \bibinfo{person}{V. Dubourg}, \bibinfo{person}{J.
  Vanderplas}, \bibinfo{person}{A. Passos}, \bibinfo{person}{D. Cournapeau},
  \bibinfo{person}{M. Brucher}, \bibinfo{person}{M. Perrot}, {and}
  \bibinfo{person}{E. Duchesnay}.} \bibinfo{year}{2011}\natexlab{}.
\newblock \showarticletitle{Scikit-learn: Machine Learning in {P}ython}.
\newblock \bibinfo{journal}{\emph{Journal of Machine Learning Research}}
  \bibinfo{volume}{12} (\bibinfo{year}{2011}), \bibinfo{pages}{2825--2830}.
\newblock


\bibitem[\protect\citeauthoryear{Peel and Clauset}{Peel and Clauset}{2015}]%
        {peel2015detecting}
\bibfield{author}{\bibinfo{person}{Leto Peel} {and} \bibinfo{person}{Aaron
  Clauset}.} \bibinfo{year}{2015}\natexlab{}.
\newblock \showarticletitle{Detecting change points in the large-scale
  structure of evolving networks}. In \bibinfo{booktitle}{\emph{Twenty-Ninth
  AAAI Conference on Artificial Intelligence}}.
\newblock


\bibitem[\protect\citeauthoryear{Ranshous, Shen, Koutra, Harenberg, Faloutsos,
  and Samatova}{Ranshous et~al\mbox{.}}{2015}]%
        {ranshous2015anomaly}
\bibfield{author}{\bibinfo{person}{Stephen Ranshous}, \bibinfo{person}{Shitian
  Shen}, \bibinfo{person}{Danai Koutra}, \bibinfo{person}{Steve Harenberg},
  \bibinfo{person}{Christos Faloutsos}, {and} \bibinfo{person}{Nagiza~F
  Samatova}.} \bibinfo{year}{2015}\natexlab{}.
\newblock \showarticletitle{Anomaly detection in dynamic networks: a survey}.
\newblock \bibinfo{journal}{\emph{Wiley Interdisciplinary Reviews:
  Computational Statistics}} \bibinfo{volume}{7}, \bibinfo{number}{3}
  (\bibinfo{year}{2015}), \bibinfo{pages}{223--247}.
\newblock


\bibitem[\protect\citeauthoryear{Rufai, Anbarjafari, and Demirel}{Rufai
  et~al\mbox{.}}{2014}]%
        {rufai2014lossy}
\bibfield{author}{\bibinfo{person}{Awwal~Mohammed Rufai},
  \bibinfo{person}{Gholamreza Anbarjafari}, {and} \bibinfo{person}{Hasan
  Demirel}.} \bibinfo{year}{2014}\natexlab{}.
\newblock \showarticletitle{Lossy image compression using singular value
  decomposition and wavelet difference reduction}.
\newblock \bibinfo{journal}{\emph{Digital signal processing}}
  \bibinfo{volume}{24} (\bibinfo{year}{2014}), \bibinfo{pages}{117--123}.
\newblock


\bibitem[\protect\citeauthoryear{Shah, Koutra, Zou, Gallagher, and
  Faloutsos}{Shah et~al\mbox{.}}{2015}]%
        {shah2015timecrunch}
\bibfield{author}{\bibinfo{person}{Neil Shah}, \bibinfo{person}{Danai Koutra},
  \bibinfo{person}{Tianmin Zou}, \bibinfo{person}{Brian Gallagher}, {and}
  \bibinfo{person}{Christos Faloutsos}.} \bibinfo{year}{2015}\natexlab{}.
\newblock \showarticletitle{Timecrunch: Interpretable dynamic graph
  summarization}. In \bibinfo{booktitle}{\emph{Proceedings of the 21th ACM
  SIGKDD International Conference on Knowledge Discovery and Data Mining}}.
  \bibinfo{pages}{1055--1064}.
\newblock


\bibitem[\protect\citeauthoryear{Shi and Malik}{Shi and Malik}{2000}]%
        {shi2000normalized}
\bibfield{author}{\bibinfo{person}{Jianbo Shi} {and} \bibinfo{person}{Jitendra
  Malik}.} \bibinfo{year}{2000}\natexlab{}.
\newblock \showarticletitle{Normalized cuts and image segmentation}.
\newblock \bibinfo{journal}{\emph{IEEE Transactions on pattern analysis and
  machine intelligence}} \bibinfo{volume}{22}, \bibinfo{number}{8}
  (\bibinfo{year}{2000}), \bibinfo{pages}{888--905}.
\newblock


\bibitem[\protect\citeauthoryear{Thompson and Eliassi-Rad}{Thompson and
  Eliassi-Rad}{2009}]%
        {thompson2009dapa}
\bibfield{author}{\bibinfo{person}{Brian Thompson} {and} \bibinfo{person}{Tina
  Eliassi-Rad}.} \bibinfo{year}{2009}\natexlab{}.
\newblock \bibinfo{booktitle}{\emph{Dapa-v10: Discovery and analysis of
  patterns and anomalies in volatile time-evolving networks}}.
\newblock \bibinfo{type}{{T}echnical {R}eport}. \bibinfo{institution}{Lawrence
  Livermore National Lab.(LLNL), Livermore, CA (United States)}.
\newblock


\bibitem[\protect\citeauthoryear{{Virtanen}, {Gommers}, {Oliphant},
  {Haberland}, {Reddy}, {Cournapeau}, {Burovski}, {Peterson}, {Weckesser},
  {Bright}, {van der Walt}, {Brett}, {Wilson}, {Jarrod Millman}, {Mayorov},
  {Nelson}, {Jones}, {Kern}, {Larson}, {Carey}, {Polat}, {Feng}, {Moore}, {Vand
  erPlas}, {Laxalde}, {Perktold}, {Cimrman}, {Henriksen}, {Quintero}, {Harris},
  {Archibald}, {Ribeiro}, {Pedregosa}, {van Mulbregt}, and
  {Contributors}}{{Virtanen} et~al\mbox{.}}{2020}]%
        {2020SciPy-NMeth}
\bibfield{author}{\bibinfo{person}{Pauli {Virtanen}}, \bibinfo{person}{Ralf
  {Gommers}}, \bibinfo{person}{Travis~E. {Oliphant}}, \bibinfo{person}{Matt
  {Haberland}}, \bibinfo{person}{Tyler {Reddy}}, \bibinfo{person}{David
  {Cournapeau}}, \bibinfo{person}{Evgeni {Burovski}}, \bibinfo{person}{Pearu
  {Peterson}}, \bibinfo{person}{Warren {Weckesser}}, \bibinfo{person}{Jonathan
  {Bright}}, \bibinfo{person}{St{\'e}fan~J. {van der Walt}},
  \bibinfo{person}{Matthew {Brett}}, \bibinfo{person}{Joshua {Wilson}},
  \bibinfo{person}{K. {Jarrod Millman}}, \bibinfo{person}{Nikolay {Mayorov}},
  \bibinfo{person}{Andrew R.~J. {Nelson}}, \bibinfo{person}{Eric {Jones}},
  \bibinfo{person}{Robert {Kern}}, \bibinfo{person}{Eric {Larson}},
  \bibinfo{person}{CJ {Carey}}, \bibinfo{person}{{\.I}lhan {Polat}},
  \bibinfo{person}{Yu {Feng}}, \bibinfo{person}{Eric~W. {Moore}},
  \bibinfo{person}{Jake {Vand erPlas}}, \bibinfo{person}{Denis {Laxalde}},
  \bibinfo{person}{Josef {Perktold}}, \bibinfo{person}{Robert {Cimrman}},
  \bibinfo{person}{Ian {Henriksen}}, \bibinfo{person}{E.~A. {Quintero}},
  \bibinfo{person}{Charles~R {Harris}}, \bibinfo{person}{Anne~M. {Archibald}},
  \bibinfo{person}{Ant{\^o}nio~H. {Ribeiro}}, \bibinfo{person}{Fabian
  {Pedregosa}}, \bibinfo{person}{Paul {van Mulbregt}}, {and}
  \bibinfo{person}{SciPy 1.~0 {Contributors}}.}
  \bibinfo{year}{2020}\natexlab{}.
\newblock \showarticletitle{{SciPy 1.0: Fundamental Algorithms for Scientific
  Computing in Python}}.
\newblock \bibinfo{journal}{\emph{Nature Methods}} (\bibinfo{year}{2020}).
\newblock
\urldef\tempurl%
\url{https://doi.org/10.1038/s41592-019-0686-2}
\showDOI{\tempurl}


\bibitem[\protect\citeauthoryear{Von~Luxburg}{Von~Luxburg}{2007}]%
        {von2007tutorial}
\bibfield{author}{\bibinfo{person}{Ulrike Von~Luxburg}.}
  \bibinfo{year}{2007}\natexlab{}.
\newblock \showarticletitle{A tutorial on spectral clustering}.
\newblock \bibinfo{journal}{\emph{Statistics and computing}}
  \bibinfo{volume}{17}, \bibinfo{number}{4} (\bibinfo{year}{2007}),
  \bibinfo{pages}{395--416}.
\newblock


\bibitem[\protect\citeauthoryear{Wang, Chakrabarti, Sivakoff, and
  Parthasarathy}{Wang et~al\mbox{.}}{2017}]%
        {wang2017fast}
\bibfield{author}{\bibinfo{person}{Yu Wang}, \bibinfo{person}{Aniket
  Chakrabarti}, \bibinfo{person}{David Sivakoff}, {and}
  \bibinfo{person}{Srinivasan Parthasarathy}.} \bibinfo{year}{2017}\natexlab{}.
\newblock \showarticletitle{Fast change point detection on dynamic social
  networks}.
\newblock \bibinfo{journal}{\emph{arXiv preprint arXiv:1705.07325}}
  (\bibinfo{year}{2017}).
\newblock


\bibitem[\protect\citeauthoryear{Xu, Hu, Leskovec, and Jegelka}{Xu
  et~al\mbox{.}}{2018}]%
        {xu2018powerful}
\bibfield{author}{\bibinfo{person}{Keyulu Xu}, \bibinfo{person}{Weihua Hu},
  \bibinfo{person}{Jure Leskovec}, {and} \bibinfo{person}{Stefanie Jegelka}.}
  \bibinfo{year}{2018}\natexlab{}.
\newblock \showarticletitle{How powerful are graph neural networks?}
\newblock \bibinfo{journal}{\emph{arXiv preprint arXiv:1810.00826}}
  (\bibinfo{year}{2018}).
\newblock


\bibitem[\protect\citeauthoryear{Yu, Qiu, Wen, Lin, and Liu}{Yu
  et~al\mbox{.}}{2016}]%
        {yu2016survey}
\bibfield{author}{\bibinfo{person}{Rose Yu}, \bibinfo{person}{Huida Qiu},
  \bibinfo{person}{Zhen Wen}, \bibinfo{person}{ChingYung Lin}, {and}
  \bibinfo{person}{Yan Liu}.} \bibinfo{year}{2016}\natexlab{}.
\newblock \showarticletitle{A survey on social media anomaly detection}.
\newblock \bibinfo{journal}{\emph{ACM SIGKDD Explorations Newsletter}}
  \bibinfo{volume}{18}, \bibinfo{number}{1} (\bibinfo{year}{2016}),
  \bibinfo{pages}{1--14}.
\newblock


\bibitem[\protect\citeauthoryear{Yu, Cheng, Aggarwal, Zhang, Chen, and Wang}{Yu
  et~al\mbox{.}}{2018}]%
        {yu2018netwalk}
\bibfield{author}{\bibinfo{person}{Wenchao Yu}, \bibinfo{person}{Wei Cheng},
  \bibinfo{person}{Charu~C Aggarwal}, \bibinfo{person}{Kai Zhang},
  \bibinfo{person}{Haifeng Chen}, {and} \bibinfo{person}{Wei Wang}.}
  \bibinfo{year}{2018}\natexlab{}.
\newblock \showarticletitle{Netwalk: A flexible deep embedding approach for
  anomaly detection in dynamic networks}. In
  \bibinfo{booktitle}{\emph{Proceedings of the 24th ACM SIGKDD International
  Conference on Knowledge Discovery \& Data Mining}}. ACM,
  \bibinfo{pages}{2672--2681}.
\newblock


\bibitem[\protect\citeauthoryear{Zamani, Nanjundaswamy, and Rose}{Zamani
  et~al\mbox{.}}{2017}]%
        {zamani2017frequency}
\bibfield{author}{\bibinfo{person}{Sina Zamani}, \bibinfo{person}{Tejaswi
  Nanjundaswamy}, {and} \bibinfo{person}{Kenneth Rose}.}
  \bibinfo{year}{2017}\natexlab{}.
\newblock \showarticletitle{Frequency domain singular value decomposition for
  efficient spatial audio coding}. In \bibinfo{booktitle}{\emph{2017 IEEE
  Workshop on Applications of Signal Processing to Audio and Acoustics
  (WASPAA)}}. IEEE, \bibinfo{pages}{126--130}.
\newblock


\bibitem[\protect\citeauthoryear{Zhang}{Zhang}{2011}]%
        {zhang2011laplacian}
\bibfield{author}{\bibinfo{person}{Xiao-Dong Zhang}.}
  \bibinfo{year}{2011}\natexlab{}.
\newblock \showarticletitle{The Laplacian eigenvalues of graphs: a survey}.
\newblock \bibinfo{journal}{\emph{arXiv preprint arXiv:1111.2897}}
  (\bibinfo{year}{2011}).
\newblock


\bibitem[\protect\citeauthoryear{Zheng, Li, Li, Li, and Gao}{Zheng
  et~al\mbox{.}}{2019}]%
        {zheng2019addgraph}
\bibfield{author}{\bibinfo{person}{Li Zheng}, \bibinfo{person}{Zhenpeng Li},
  \bibinfo{person}{Jian Li}, \bibinfo{person}{Zhao Li}, {and}
  \bibinfo{person}{Jun Gao}.} \bibinfo{year}{2019}\natexlab{}.
\newblock \showarticletitle{Addgraph: anomaly detection in dynamic graph using
  attention-based temporal GCN}. In \bibinfo{booktitle}{\emph{Proceedings of
  the 28th International Joint Conference on Artificial Intelligence}}. AAAI
  Press, \bibinfo{pages}{4419--4425}.
\newblock


\bibitem[\protect\citeauthoryear{Ziegel}{Ziegel}{2001}]%
        {ziegel2001standard}
\bibfield{author}{\bibinfo{person}{Eric~R Ziegel}.}
  \bibinfo{year}{2001}\natexlab{}.
\newblock \showarticletitle{Standard probability and statistics tables and
  formulae}.
\newblock \bibinfo{journal}{\emph{Technometrics}} \bibinfo{volume}{43},
  \bibinfo{number}{2} (\bibinfo{year}{2001}), \bibinfo{pages}{249}.
\newblock


\end{thebibliography}

\newpage
\appendix

\section{Reproducibility}
\label{app:repro}

We report the implementations used in our experiments. 
For TENSORSPLAT, we first compute the PARAFAC decomposition~(using Tensorly~\cite{kossaifi2019tensorly} library in python) to obtain the temporal factors. Then the scikit-learn~\cite{scikit-learn} python implementation of the Local Outlier Factor algorithm. For EdgeMonitoring method, we use the matlab code kindly provided by the original authors and keep the default parameters.

\section{Spectral Properties and Their Connections}
\label{app:spec}

\begin{table}[ht]
\centering
\begin{tabular}{c|c}
\hline
\hline
Spectral Properties & Connections \\
\hline 
$\mathbf{L},\mathbf{L}_{rw},\mathbf{L}_{sym}$ eigenvalues & connected components~\cite{von2007tutorial} \\

$\mathbf{L}$ eigenvectors & ratio cut~\cite{hagen1992new} \\

$\mathbf{L}_{rw},\mathbf{L}_{sym}$ eigenvectors & normalized cut~\cite{shi2000normalized,von2007tutorial} \\

$\mathbf{A}$ eigenvalues & KATZ centrality~\cite{goh2001universal}\\
$\mathbf{A}$ eigenvectors & eigenvector centrality~\cite{bonacich1987power}\\
$\mathbf{A}$ dominant eigenvector &stationary distrib., PageRank~\cite{page1999pagerank}\\
\hline
\end{tabular}
\caption{Spectral Properties and Their Connections}
\label{tab:connections}
\end{table}

The above table summarizes connections between different spectral properties in the graph and their connections in the literature. We use the same notation as~\cite{von2007tutorial}. $\mathbf{L},\mathbf{L}_{rw},\mathbf{L}_{sym}, \mathbf{A}$ represent the unnormalized Laplacian matrix, the random walk Laplacian matrix, the symmetric Laplacian matrix and the adjacency matrix respectively.

\section{Canadian bill voting network}
\label{app:canadian}

The Canadian bill voting network was mined from the Open Parliament API~(http://api.openparliament.ca/), a source for digitized data from the House of Commons in JSON format. We first extracted all MPs in the canadian parliament from 2006 to 2019. The network nodes for each snapshot only includes the MPs who actively participated in the parliament of that year~(around 300 MPs depending on which year). We then extracted all bills sponsored by each MP and the corresponding votes for each bill. Lastly, we filtered out the yes votes for each ballot of the bills and which MPs voted yes. The data mining code is also available in the code repository of the project~\footnote{\url{https://github.com/shenyangHuang/LAD}}.

We also provide more information on the political environment from 2006 to 2019. During this time period the government party in power has changed. In 2006, the government was minority Conservative until 2015 when the Liberal party won and formed a majority parliament~\cite{macfarlane2019renewed}. Studies have shown that minority governments appear to be less productive in legislative activity as consensus is harder to obtain~\cite{conley2011legislative}. Cohesion in the House of Commons within a political party during voting sessions are often observed and dissent has been seen amongst MPs of the same party who are less influential~\cite{garner2005party}. While elected to the House of Commons, MPs can sponsor more than one bill which can also include bills that may have not passed in prior parliaments. Parliament sessions follows no regular pattern from one parliament to another.

\end{document}